  \documentclass[10pt,twocolumn]{article}
\usepackage{hmi-preprint}
\usepackage{amsmath}
\usepackage{amsfonts}
\usepackage{url}
\usepackage{booktabs}
\usepackage{multirow}
\usepackage{array}
\usepackage{makecell}
\usepackage{enumitem}
\usepackage{float}
\usepackage{balance}
\usepackage{nicefrac}
\usepackage{microtype}
\usepackage{marvosym}
\usepackage{xcolor}         % colors
\definecolor{cvprblue}{rgb}{0.21,0.49,0.74}
\hypersetup{
    colorlinks=true, % 开启链接颜色（将方框改为彩色文字）
    linkcolor=cvprblue, % 内部链接颜色（如目录、图表引用）
    citecolor=cvprblue, % 文献引用颜色
    urlcolor=cvprblue,  % 外部网页链接颜色
}

\preprintlogos{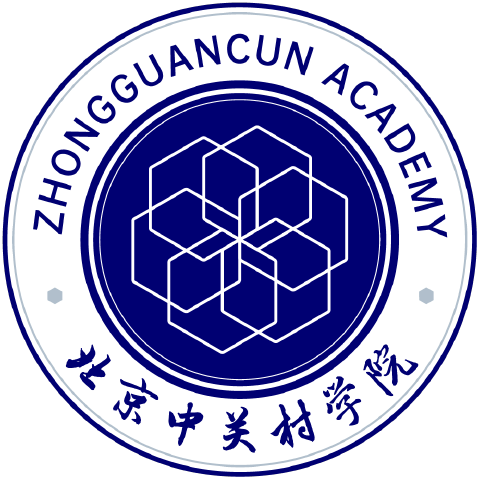}{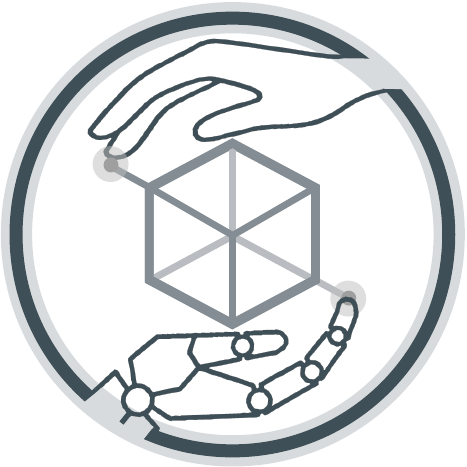}

\preprintdate{May 2026}

\runningtitle{GeoHand: Unlocking Prior Geometry Knowledge for Monocular 3D Hand Reconstruction}

\preprintfooter{\textsuperscript{\Letter} Corresponding author.}

\preprintlinkset{}{}{}

\newcommand{\GeoHandFigWidth}{1.0\textwidth}
\newcommand{\VisualizationFig}{0.9\textwidth}
\newcommand{\VisualizationFigTwo}{0.8\textwidth}
\DeclareMathOperator{\sigmoid}{sigmoid}
\DeclareMathOperator{\MLP}{MLP}
\DeclareMathOperator{\LN}{LayerNorm}
\DeclareMathOperator{\Proj}{Proj}
\DeclareMathOperator{\Concat}{Concat}

\preprintteaser{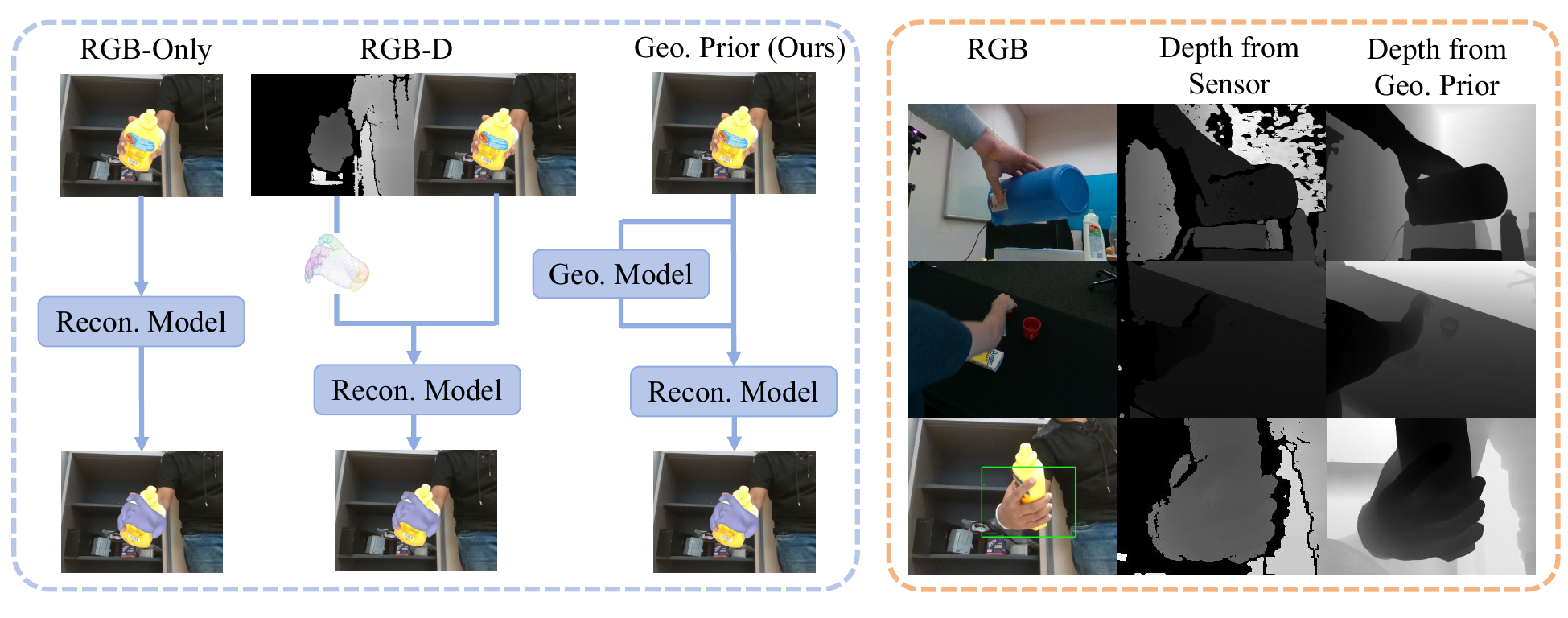}{Comparison of 3D reconstruction pipelines. Unlike pure RGB or RGB-D methods, GeoHand extracts high-fidelity spatial priors from a foundational monocular geometry model to recover accurate 3D hands under severe occlusions. The right panel contrasts depth maps from localized hand crops (third row), the actual input format for GeoHand, highlighting that our prior yields noise-free, smoothly interpolated depth even on tight crops, bypassing missing values typical of raw sensor data.}

\title{GeoHand: Unlocking Prior Geometry Knowledge for Monocular 3D Hand Reconstruction}

\author{
Weiquan Lin\textsuperscript{1,2} \hspace{0.12in}
Yaoqing Hu\textsuperscript{2} \hspace{0.12in}
Liangchen Dai\textsuperscript{3,2} \hspace{0.12in}
Xu Tang\textsuperscript{1} \hspace{0.12in}
Xingyu Chen\textsuperscript{2 \Letter}
}

\affil{
\textsuperscript{1}School of Artificial Intelligence, Xidian University\\
\textsuperscript{2}Zhongguancun Academy\\
\textsuperscript{3}School of Automation, Beijing Institute of Technology
}

\begin{document}

\maketitle

\begin{abstract}

Monocular 3D hand reconstruction is intrinsically a geometric problem, yet RGB appearance features alone often struggle to resolve severe ambiguities caused by self-occlusions and hand-object interactions. While introducing depth can explicitly provide spatial cues, raw sensor-captured depth maps are extensively noisy and incomplete, limiting their usefulness for fine-grained hand reconstruction. To bridge this gap, we propose GeoHand, a novel framework that unlocks high-quality geometric priors from a frozen foundational monocular geometry estimator (MoGe2). Recognizing that these priors are oriented toward general scenes, we introduce a map-level GeoAdapter to recalibrate the spatial features, specifically adapting them for detailed hand reconstruction. Furthermore, to systematically integrate these adapted priors without overwhelming intrinsic RGB appearance cues, we employ a gated cross-modal token fusion strategy. Finally, to secure precise local articulation, we design a Keypoint-Queried Iterative Refiner (KQIR) that uses projected joint locations to query geometry-aware image features for spatial correction. By combining global geometric disambiguation with local refinement in a unified pipeline, GeoHand achieves state-of-the-art performance on FreiHAND, DexYCB, and HO3Dv3, especially under severe occlusions and hand-object interactions.
\end{abstract}

\section{Introduction}
\label{sec:intro}

Recovering the 3D pose and shape of human hands from a single RGB image is a fundamental and longstanding challenge in computer vision. Accurate 3D hand reconstruction serves as the cornerstone for numerous downstream applications, ranging from immersive augmented and virtual reality (AR/VR) experiences to sophisticated human-computer interaction (HCI) and embodied robotics. Unlike rigid objects, the human hand is highly articulated with numerous degrees of freedom. Consequently, estimating 3D hand parameters from a 2D projection is intrinsically ill-posed. A single RGB image provides minimal evidence regarding metric depth, global scale, and the complex topological relationships of self-occluded fingers \cite{zimmermann2019freihand, chen2022mobrecon, pavlakos2024reconstructing, pavlakos2019expressive}. This inherent depth ambiguity and widespread self-occlusion make the precise reconstruction of 3D hand meshes a substantially more formidable task than 2D keypoint localization.

A dominant paradigm in recent years has focused on improving monocular reconstruction by scaling up the capacity of Transformer backbones, designing more sophisticated decoders, or leveraging massive amounts of training data. Early pioneering approaches attempted to directly regress the parametric MANO model \cite{romero2022embodied, bogo2016keep, kolotouros2019learning, oberweger2015hands} using convolutional neural networks \cite{boukhayma20193d, zhang2019end, huang2020awr} or large-scale synthetic data generation \cite{mueller2018ganerated}. To capture the complex non-linear mapping between 2D pixels and 3D meshes, subsequent methods introduced graph convolutional networks \cite{kulon2020weakly, baek2019pushing, ge20193d, kolotouros2019convolutional} and attention-based architectures \cite{chen2021camera}. Most recently, the advent of vision transformers (ViTs) \cite{dosovitskiy2020image, he2022masked, liu2021swin} has sparked a paradigm shift. Models such as HaMeR \cite{pavlakos2024reconstructing} and WiLoR \cite{potamias2025wilor} utilize large-scale transformer backbones to capture long-range visual dependencies \cite{lin2021end, lin2021mesh, cho2022cross}, while SimpleHand \cite{zhou2024simple} shows that carefully designed token generation can achieve a strong accuracy--efficiency trade-off. Despite these impressive algorithmic strides, the core bottleneck remains: these methods force the backbone to infer 3D geometry implicitly from 2D appearance cues alone.

To mitigate the geometric ambiguity of RGB-only inputs, an alternative line of research explores multi-modal or geometry-aware reasoning. Utilizing auxiliary signals, such as depth maps or IMU measurements, has historically been the most direct way to resolve scale and occlusion ambiguities \cite{moon2018v2v, chao2021dexycb, hampali2020honnotate}. However, explicit multi-modal pipelines face two practical challenges. First, they heavily depend on the availability of paired multi-modal datasets (e.g., synchronized RGB-D), whose large-scale acquisition is restricted by high construction costs \cite{taheri2020grab, hasson2019learning, brahmbhatt2020contactpose}. Furthermore, requiring explicit multi-modal inputs during inference limits general deployment. Second, their fusion strategies \cite{liu2024keypoint, qi2017pointnet, qi2017pointnet++} typically entail highly intricate point-cloud or volumetric alignment, which complicates optimization. The recent evolution of foundational monocular depth and geometry estimators provides a promising avenue to bypass these hurdles. General-purpose models such as Depth Anything \cite{yang2024depth} and MoGe2 \cite{wang2025moge} exhibit a remarkable ability to extract dense, metric-scale 3D geometric structures directly from a single RGB image. Crucially, as illustrated in Figure \ref{fig:teaser}, the geometric representations of MoGe2 exhibit pronounced advantages over raw depth sensors even in localized hand crop scenarios. 
% While these foundational models are typically trained on full uncropped scenes, we demonstrate that their extracted spatial bounds remain tightly reliable even when applied directly to localized bounding-box hand crops. 
This capability raises a critical question: \textit{Can we seamlessly inject these powerful, frozen geometric priors into a hand reconstruction pipeline without compromising the simplicity of RGB-only inference?}

In this paper, we answer this question by proposing GeoHand, a framework that unlocks prior geometry knowledge for monocular 3D hand reconstruction. GeoHand preserves the standard single-RGB input pipeline but systematically enriches the internal representation using a frozen monocular geometry model (MoGe2). Recognizing that naively concatenating geometry tokens often leads to feature misalignment, we introduce a lightweight, map-level GeoAdapter. This module adaptively recalibrates intermediate geometry feature maps before tokenization. The resulting geometry tokens are then integrated into the primary transformer backbone via a gated cross-modal token fusion strategy. 

A key challenge remains, however: global geometric priors can reduce coarse depth ambiguity, but reconstruction errors in hand meshes are often concentrated at highly articulated local structures such as fingertips and inter-finger occlusions. To address this, we introduce a Keypoint-Queried Iterative Refiner (KQIR). Rather than acting as an isolated post-processing block, KQIR explicitly builds on the geometry-aware image representation produced by GeoAdapter and token fusion. It uses the current 3D joints as structured queries to retrieve local evidence from geometry-aware features, thereby translating global geometric disambiguation into targeted local articulation correction.

In summary, our contributions are summarized as follows:
\begin{itemize}[leftmargin=1.5em]
    \item We propose GeoHand, a novel monocular hand reconstruction framework that effectively unlocks prior geometry knowledge, bridging the gap between RGB-based appearance features and foundational spatial priors.
    \item We introduce GeoAdapter with a gated token fusion mechanism to adaptively extract and align frozen geometry features, together with a lightweight Keypoint-Queried Iterative Refiner (KQIR) that exploits these geometry-aware features for local joint refinement.
    \item Extensive evaluations on FreiHAND, DexYCB, and HO3Dv3 demonstrate that GeoHand achieves state-of-the-art reconstruction performance. In particular, our model excels under heavy object-induced occlusion, despite using an order of magnitude less annotated data than concurrent massive-scale paradigms on the challenging HO3Dv3 benchmark.
\end{itemize}

\section{Related Work}
\label{sec:related}

\subsection{Monocular RGB Hand Reconstruction}
The pursuit of estimating 3D hand posture from a single RGB image has historically been divided into model-free and model-based approaches. While model-free methods regress 3D joint coordinates directly \cite{zimmermann2017learning, iqbal2018hand, cai2018weakly, ge2018hand}, model-based works regress parametric models like MANO \cite{romero2022embodied} to ensure anatomical plausibility \cite{boukhayma20193d, kanazawa2018end, rong2021frankmocap}. Architectures have evolved from ResNets to graph convolutional networks \cite{ge20193d, kulon2020weakly, tse2022collaborative, choi2020pose2mesh} and disentangled representation learning frameworks \cite{yang2019disentangling}. Other techniques explore dense pixel-to-surface mappings \cite{guler2018densepose, moon2020i2l} or probabilistic continuous distributions \cite{kolotouros2021probabilistic}. Recently, models such as MobRecon \cite{chen2022mobrecon}, HandOccNet \cite{park2022handoccnet}, H2ONet \cite{xu2023h2onet}, HaMeR \cite{pavlakos2024reconstructing}, and WiLoR \cite{potamias2025wilor} leverage ViTs to establish long-range spatial correlations, significantly improving robustness in highly occluded in-the-wild images. However, these RGB-only transformers fundamentally rely on appearance cues to infer 3D depth variations implicitly for hand pose estimation.

\subsection{Multi-Modal and Geometry-Aware Methods}
To alleviate the inherent ambiguity of RGB imagery, incorporating explicit 3D spatial information has proven highly effective. Depth maps provided by RGB-D sensors offer direct geometric observations \cite{moon2018v2v, garcia2018first}. In hand-object interaction scenarios, leveraging object geometry as an explicit constraint has been thoroughly explored in DexYCB \cite{chao2021dexycb} and HO3Dv3 \cite{hampali2020honnotate}, along with large interacting-hand datasets such as InterHand2.6M \cite{moon2020interhand2}, H2O \cite{kwon2021h2o}, OakInk \cite{yang2022oakink}, and DexGraspNet \cite{wang2022dexgraspnet}. Multi-modal fusion frameworks have utilized diverse signals spanning physical contacts \cite{grady2021contactopt, karunratanakul2020grasping}, point clouds \cite{liu2024keypoint, li2022interacting}, event cameras \cite{jiang2024complementing}, and cross-modal autoencoders \cite{spurr2018cross}.

% However, these systems often heavily depend on the availability of paired multi-modal annotations during training, which limits algorithmic generalizability because such high-quality hardware-captured datasets are difficult to construct at scale. Explicit multi-modal inputs during inference further necessitate calibrated auxiliary sensor deployment. Additionally, their explicit fusion strategies are often algorithmically intricate and unstructured. GeoHand circumvents both the heavy reliance on paired multi-modal data collection and the complexity of explicit cross-modal alignment. By extracting high-fidelity geometric priors entirely from monocular RGB images, GeoHand unlocks structural priors without sacrificing the accessibility of RGB-only inference.

However, these systems heavily depend on the availability of paired multi-modal annotations during training. This dependence limits algorithmic generalizability, as large-scale hardware-captured datasets are notoriously difficult to construct. Furthermore, as illustrated in Figure \ref{fig:teaser}, raw sensor depth frequently suffers from noise and missing values, which undermines its reliability for reconstructing fine-grained hand structures. At inference time, requiring explicit multi-modal inputs also necessitates calibrated auxiliary sensor deployment, while their explicit fusion strategies are often algorithmically intricate and unstructured. GeoHand circumvents both the rigorous demands of multi-modal data collection and the complexity of cross-modal alignment. By extracting high-fidelity geometric priors entirely from monocular RGB images, GeoHand unlocks robust structural guidance without sacrificing the flexibility and accessibility of RGB-only inference.

\subsection{Monocular Geometry Priors for 3D Vision}
Estimating depth and surface normals from a single RGB image is a cornerstone of scene understanding \cite{ranftl2020towards}. This paradigm has recently expanded rapidly with foundational models like Depth Anything \cite{yang2024depth}, diffusion-based estimators like Marigold \cite{ke2024repurposing}, and general spatial representation models like MoGe \cite{wang2024moge} and MoGe2 \cite{wang2025moge}. While these priors have revolutionized broad scene reconstruction, their application to articulated hand mesh recovery remains underexplored. Naively appending depth maps typically yields suboptimal results due to domain gaps between general scene scales and fine-grained hand articulations \cite{zimmermann2021contrastive, yang2020bihand}. GeoHand addresses this gap by moving beyond pixel-level depth fusion. Instead, we extract intermediate structural features from a frozen MoGe2 backbone and adapt them using GeoAdapter. By employing lightweight convolutional recalibration, GeoAdapter explicitly isolates and aligns hand-specific geometric structures from general backgrounds. The adapted semantic geometry features are then systematically fused at the compact 1D token level.

\section{Method}
\label{sec:method}

\subsection{Overview of the Architecture}
\label{sec:overview}

Given a cropped RGB image $\mathbf{I}\in\mathbb{R}^{3\times H\times W}$, GeoHand predicts MANO pose parameters $\boldsymbol{\theta}$, shape parameters $\boldsymbol{\beta}$, and weak-perspective camera parameters $\mathbf{c}$, which are then decoded into 3D joints $\mathbf{J}$ and mesh vertices $\mathbf{V}$. In our implementation, the input is resized to $256\times192$, yielding a $16\times12$ patch grid. Specifically, the RGB branch uses an embedding module in a ViT-L transformer to extract spatial tokens:
\begin{equation}
\mathbf{X}_{\mathrm{rgb}} = \text{PatchEmbed}(\mathbf{I}) + \mathbf{P}_{\mathrm{rgb}} \in \mathbb{R}^{B\times N\times D},
\end{equation}
where $B$ is the batch size, $N$ is the number of patch tokens, and $D$ is the token dimension.

\begin{figure*}[t!]
\centering
\includegraphics[width=\GeoHandFigWidth,keepaspectratio]{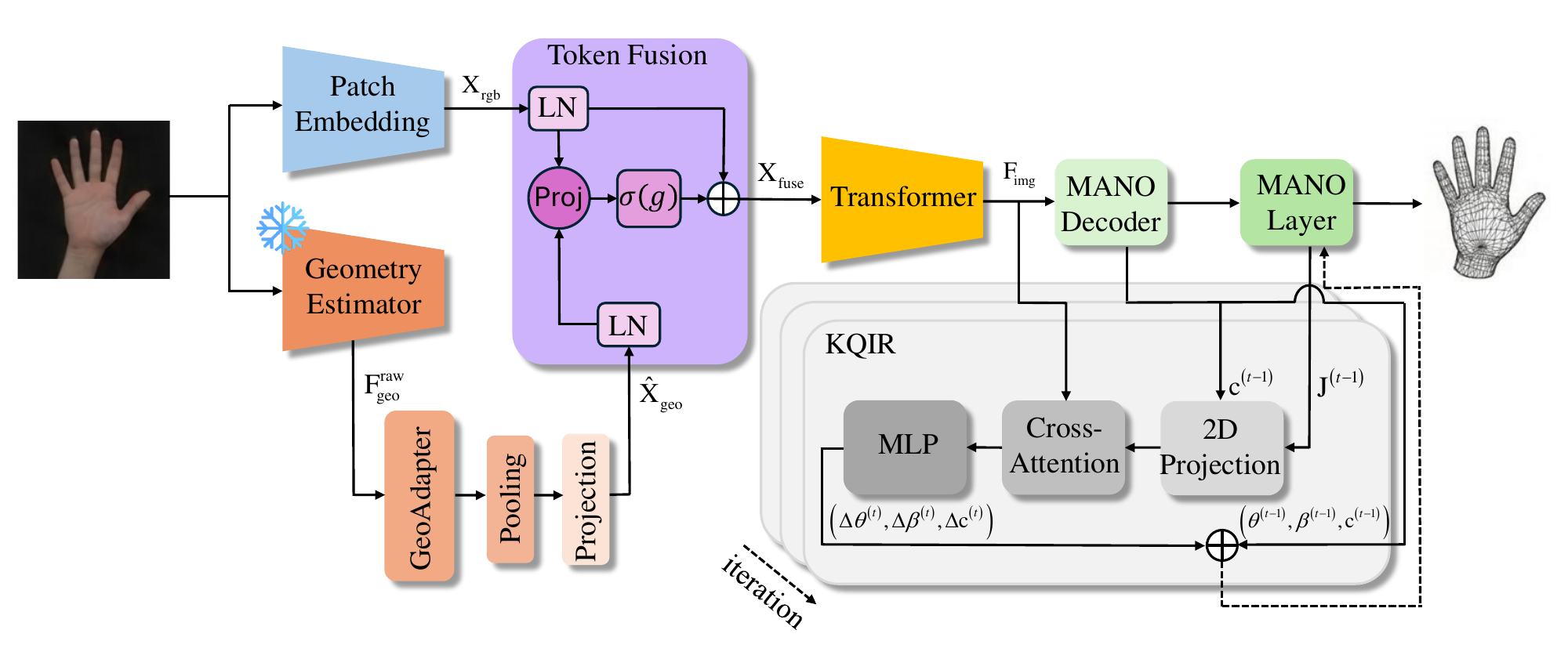}
\caption{Overall pipeline of GeoHand. The network extracts parallel appearance tokens and prior geometry features (adapted via GeoAdapter). These are integrated using a gated cross-modal token fusion strategy. Finally, a MANO decoder predicts a coarse mesh, which is subsequently refined using the Keypoint-Queried Iterative Refiner (KQIR).}
\label{fig:geohand_framework}
\end{figure*}

As shown in Fig.~\ref{fig:geohand_framework}, GeoHand computes visual features in parallel with a frozen MoGe2 geometry estimator. Given the processed image $\tilde{\mathbf{I}}$, we extract an intermediate geometry feature map directly from the foundational geometry model:
\begin{equation}
\mathbf{F}_{\mathrm{geo}}^{\mathrm{raw}} = \mathcal{G}(\tilde{\mathbf{I}}) \in \mathbb{R}^{B\times C_g\times H_g\times W_g},
\end{equation}
where $\mathcal{G}$ denotes the frozen geometry network and $C_g$ is its output channel dimension. These two streams establish the appearance and geometry representations required for downstream alignment.

\subsection{Geometry Tokenization via GeoAdapter}

Raw geometry features from general scenes exhibit a domain gap when applied directly to localized finger structures. We therefore employ a spatial-domain GeoAdapter to recalibrate these geometry features before they are collapsed into 1D sequences. Using a lightweight stack of $1\times1$ and $3\times3$ convolutions (with GroupNorm and GELU), the adapter predicts an augmenting side feature:
\begin{equation}
\mathbf{F}_{\mathrm{side}} = \mathcal{A}(\mathbf{F}_{\mathrm{geo}}^{\mathrm{raw}}) \in \mathbb{R}^{B\times C_s\times H_g\times W_g}.
\end{equation}
This side feature is concatenated with the raw map along the channel dimension to form the adapted representation:
\begin{equation}
\mathbf{F}_{\mathrm{geo}}^{\mathrm{aug}} = \Concat\!\left[\mathbf{F}_{\mathrm{geo}}^{\mathrm{raw}}, \mathbf{F}_{\mathrm{side}}\right].
\end{equation}

The augmented geometry map is then aligned to the ViT feature layout: it is pooled to the RGB patch grid $(H_p, W_p)$ and projected to the hidden token dimension $D$. After flattening and adding learnable grid-coordinate embeddings $\mathbf{P}_{\mathrm{geo}}$, we obtain:
\begin{equation}
\hat{\mathbf{X}}_{\mathrm{geo}} = \text{Flatten}(\text{Conv}_{1\times1}(\text{Pool}(\mathbf{F}_{\mathrm{geo}}^{\mathrm{aug}}))) + \mathbf{P}_{\mathrm{geo}}.
\end{equation}

\subsection{Gated Cross-Modal Token Fusion}
\label{sec:fusion}

Directly merging geometry and RGB tokens can degrade performance when the geometry model is uncertain under motion blur or out-of-distribution conditions. We therefore introduce a gated residual token fusion layer. For each token, we first compute the residual shift by projecting the concatenated RGB and geometry tokens back into the appearance token space:
\begin{equation}
\Delta\mathbf{x}^{(i)} = \Proj\!\left([\LN(\mathbf{x}_{\mathrm{rgb}}^{(i)}); \LN(\hat{\mathbf{x}}_{\mathrm{geo}}^{(i)})]\right).
\end{equation}
where $\Proj$ is implemented as a two-layer Multi-Layer Perceptron (MLP) mapping the concatenated features back into the appearance token space constraint. We then apply a learnable gate $g$ to control how much geometry information is injected:
\begin{equation}
\mathbf{x}_{\mathrm{fuse}}^{(i)} = \mathbf{x}_{\mathrm{rgb}}^{(i)} + \sigmoid(g)\cdot \Delta\mathbf{x}^{(i)}.
\end{equation}

The fused tokens $\mathbf{X}_{\mathrm{fuse}}$ are processed by the remaining backbone blocks $\mathcal{B}$ and reshaped into a geometry-aware image feature map:
\begin{equation}
\mathbf{F}_{\mathrm{img}} = \text{Reshape}(\mathcal{B}(\mathbf{X}_{\mathrm{fuse}})) \in \mathbb{R}^{B\times D\times H_p\times W_p}.
\end{equation}
This feature map is the unified geometry-aware representation used by the downstream decoder and refiner.

\subsection{MANO Decoding with KQIR Refinement}
\label{sec:manohead_kqir}

The feature map $\mathbf{F}_{\mathrm{img}}$ is first consumed by a transformer-based MANO decoder, which predicts a coarse hand estimate:
\begin{equation}
(\mathbf{V}^{(0)}, \mathbf{J}^{(0)}) = \text{MANO}(\boldsymbol{\theta}^{(0)}, \boldsymbol{\beta}^{(0)}).
\end{equation}

While token fusion largely resolves global depth ambiguity, errors often persist in highly articulated local regions, such as self-occluded fingertips and complex contact boundaries. Because 2D appearance in these areas is heavily obscured, the spatial estimations of coarse joints often exhibit high positional uncertainty. To recover precise local articulations, it is essential to iteratively correct the geometric evidence surrounding these uncertain areas. We therefore introduce the Keypoint-Queried Iterative Refiner (KQIR), which explicitly utilizes the current coarse 3D joint predictions as localized queries to retrieve fine-grained spatial evidence from the unified geometry-aware feature map.

At refinement step $t$, the current 3D joints are projected onto the 2D plane as $\mathbf{U}^{(t-1)} = \Pi(\mathbf{J}^{(t-1)},\mathbf{c}^{(t-1)})$. For each joint $j$, we construct a structured query using its current 3D position and projected 2D coordinate:
\begin{equation}
\mathbf{q}_{j}^{(t)} = \MLP\!\left( [\mathbf{J}_{j}^{(t-1)};\mathbf{U}_{j}^{(t-1)}] \right) \in \mathbb{R}^{d_q}.
\end{equation}
Stacking all joints yields the query matrix $\mathbf{Q}_{\mathrm{joint}}^{(t)} \in \mathbb{R}^{B\times 21\times d_q}$. To apply cross-attention, we flatten the image feature map into tokens $\mathbf{X}_{\mathrm{img}} = \text{Flatten}(\mathbf{F}_{\mathrm{img}}) \in \mathbb{R}^{B\times N\times D}$ and linearly project them to keys and values:
\begin{equation}
\mathbf{K}_{\mathrm{img}} = \mathbf{X}_{\mathrm{img}}\mathbf{W}_K, \quad \mathbf{V}_{\mathrm{img}} = \mathbf{X}_{\mathrm{img}}\mathbf{W}_V,
\end{equation}
where $\mathbf{W}_K,\mathbf{W}_V$ are learnable projection matrices. Cross-attention between the structured joint queries and image tokens is then computed as:
\begin{equation}
\mathbf{H}^{(t)} = \text{CrossAttn}(\mathbf{Q}_{\mathrm{joint}}^{(t)}, \mathbf{K}_{\mathrm{img}}, \mathbf{V}_{\mathrm{img}}).
\end{equation}
The attended joint features predict residual parameter updates via an MLP:
\begin{equation}
(\Delta\boldsymbol{\theta}^{(t)}, \Delta\boldsymbol{\beta}^{(t)}, \Delta\mathbf{c}^{(t)}) = \text{MLP}_{\mathrm{ref}}(\mathbf{H}^{(t)}).
\end{equation}
By adding these predicted residuals to the estimates from the previous step $(t-1)$, we obtain the updated parameters $(\boldsymbol{\theta}^{(t)}, \boldsymbol{\beta}^{(t)}, \mathbf{c}^{(t)})$. These parameters are then directly decoded by MANO into the refined geometry:
\begin{equation}
(\mathbf{V}^{(t)},\mathbf{J}^{(t)}) = \text{MANO}(\boldsymbol{\theta}^{(t)},\boldsymbol{\beta}^{(t)}).
\end{equation}
After $T$ refinement steps, the final prediction is $(\mathbf{V},\mathbf{J}) = (\mathbf{V}^{(T)},\mathbf{J}^{(T)})$.

In essence, GeoAdapter and token fusion inject prior geometry to reduce global ambiguity, while KQIR uses the resulting geometry-aware features to refine local joint configurations precisely where errors persist.

\subsection{Training Objectives}
\label{sec:training_objectives}

Recent top-performing frameworks, such as HaMeR \cite{pavlakos2024reconstructing} and WiLoR \cite{potamias2025wilor}, typically rely on adversarial training by introducing an auxiliary discriminator network to ensure that the predicted MANO parameters form an anatomically plausible hand. However, adversarial optimization on 3D manifolds is notoriously unstable, requires intensive hyperparameter tuning, and heavily complicates the training process. GeoHand deliberately bypasses this adversarial paradigm. Instead, we constrain the geometric solution space and explicitly enforce physical plausibility through deterministic structural and parametric loss formulations (e.g., explicit bone-length and dense vertex constraints). This strategy heavily reduces the optimization difficulty while still guarding against irregular articulations.

The overall training objective is a weighted summation of eight carefully designed loss components. These components are conceptually grouped into three categories: 2D image observation alignment, 3D structural constraints, and MANO parameter regularizations:

\begin{equation}
\begin{aligned}
\mathcal{L}_{total} &= \lambda_{2D}\mathcal{L}_{2D} + \lambda_{3D\_joint}\mathcal{L}_{3D\_joint} + \lambda_{bone}\mathcal{L}_{bone}  \\
&\quad +\lambda_{vert}\mathcal{L}_{vert} + \lambda_{global}\mathcal{L}_{global} \\
&\quad + \lambda_{pose}\mathcal{L}_{pose} + \lambda_{betas}\mathcal{L}_{betas} + \lambda_{shape}\mathcal{L}_{shape}.
\end{aligned}
\end{equation}

% \textbf{2D Image Alignment.} 
\paragraph{2D Image Alignment.}
To strictly align the reconstructed hand with image observations, $\mathcal{L}_{2D}$ calculates the $L_1$ distance between the projected 3D joints (via the predicted camera) and the ground-truth 2D coordinates.

% \textbf{3D Structural Constraints.} 
\paragraph{3D Structural Constraints.}
% Acting as an explicit geometric prior to replace an implicit discriminator, this group enforces physical validity. $\mathcal{L}_{3D\_joint}$ computes root-relative 3D joint distances, assigning $2.5\times$ weight to fingertips for challenging terminal articulations. $\mathcal{L}_{bone}$ constrains the 20 skeletal segment lengths to prevent unrealistic stretching. $\mathcal{L}_{vert}$ extends structural supervision to the dense surface by minimizing the $L_1$ distance on root-aligned MANO vertices.
Acting as an explicit geometric prior, this group enforces physical validity. $\mathcal{L}_{3D\_joint}$ computes root-relative 3D joint distances, assigning $2.5\times$ weight to fingertips for challenging terminal articulations. $\mathcal{L}_{bone}$ constrains skeletal segment lengths to prevent unrealistic stretching. $\mathcal{L}_{vert}$ extends structural supervision to the dense surface by minimizing $L_1$ distance on root-aligned MANO vertices.

% \textbf{MANO Parameter Regularizations.} 
\paragraph{MANO Parameter Regularizations.} 
To stabilize outputs within the MANO space, $\mathcal{L}_{global}$, $\mathcal{L}_{pose}$, and $\mathcal{L}_{betas}$ align the predicted global orientation, pose rotations, and shape betas with the ground truth via a Smooth-$L_1$ loss. Crucially, $\mathcal{L}_{shape}$ regularizes the $L_2$ norm of the shape vector $\boldsymbol{\beta}$ to actively prevent unbounded amplitude inflation and out-of-distribution hand thickness.

Detailed mathematical formulations and the exact empirical weights for all eight components are provided in the Appendix. 
% \ref{sec:appendix_loss}.

\section{Experiments}
\label{sec:experiments}

\subsection{Datasets and Metrics}

We evaluate GeoHand on FreiHAND \cite{zimmermann2019freihand}, DexYCB \cite{chao2021dexycb}, and HO3Dv3 \cite{hampali2020honnotate}. These benchmarks cover monocular hand reconstruction, severe object occlusion, and complex hand-object interactions. Following standard practices, we report standard benchmark metrics including PA-MPJPE, MPJPE, PA-MPVPE, MPVPE, and F-scores (Detailed in the Appendix). 
% \ref{sec:appendix_metrics}).

\subsection{Implementation Details}

The geometry branch in GeoHand extracts frozen MoGe2 neck features, which are enhanced by GeoAdapter (128 side channels, depth 2) before tokenization. The GeoTokenizer pools the augmented geometry map to a $16\times12$ layout and projects it to 1280-dimensional tokens. The MANO decoder uses 6 layers and 3 IEF iterations, followed by a KQIR refiner with 2 steps. Models are trained with AdamW \cite{loshchilov2017decoupled} using PyTorch \cite{paszke2019pytorch} for 60 epochs (base LR $2\times10^{-5}$, batch size 64).

During inference, we employ the pre-trained hand detector from WiLoR \cite{potamias2025wilor} to extract bounding boxes. It is critical to highlight the vast differences in training data scale across compared methods. Recent state-of-the-art models including HaMeR \cite{pavlakos2024reconstructing}, WiLoR \cite{potamias2025wilor}, Hamba \cite{dong2024hamba}, and HandOS \cite{chen2025handos} all utilize a massive unified dataset comprising over 2.7 million images curated by HaMeR. In contrast, our mixed-data model, denoted as GeoHand$^*$, is trained solely on a smaller compilation of Dex-YCB, HO3Dv3, and FreiHAND. For the challenging HO3Dv3 evaluation, this constitutes an order of magnitude less data ($\sim$10$\times$ fewer samples), yet GeoHand achieves superior robustness, underscoring the data efficiency unlocked by geometry priors.

\subsection{Comparison with State of the Art}

Tables \ref{tab:dexycb}, \ref{tab:ho3d}, and \ref{tab:freihand} present quantitative comparisons between GeoHand and recent leading frameworks, including MobRecon \cite{chen2022mobrecon}, SimpleHand \cite{zhou2024simple}, HaMeR \cite{pavlakos2024reconstructing}, WiLoR \cite{potamias2025wilor}, MaskHand \cite{saleem2025maskhand}, and HandOS \cite{chen2025handos}.

\begin{table}[t!]
\centering
\caption{Comparison on DexYCB.}
\label{tab:dexycb}
\setlength{\tabcolsep}{2.4pt}
\renewcommand{\arraystretch}{1.08}
\begin{tabular}{@{}lcccc@{}}
\toprule
Method & \makecell{PA-\\MPJPE$\downarrow$} & MPJPE$\downarrow$ & \makecell{PA-\\MPVPE$\downarrow$} & MPVPE$\downarrow$ \\
\midrule
MobRecon \cite{chen2022mobrecon} & 6.4 & 14.2 & 5.6 & 13.1 \\
HandOccNet \cite{park2022handoccnet} & 5.8 & 14.0 & 5.5 & 13.1 \\
SimpleHand \cite{zhou2024simple} & 5.5 & 12.4 & 5.5 & 12.1 \\
MaskHand \cite{saleem2025maskhand} & \textbf{5.0} & 11.7 & 4.9 & 11.2 \\
HandOS \cite{chen2025handos} & 5.2 & -- & 5.0 & -- \\
\midrule
GeoHand (Ours) & \textbf{5.0} & \textbf{7.4} & \textbf{4.7} & \textbf{7.1} \\
\bottomrule
\end{tabular}
\end{table}

\begin{table}[t!]
\centering
\caption{Comparison on HO3Dv3. $*$ denotes using mixed training data}
\label{tab:ho3d}
\setlength{\tabcolsep}{2.6pt}
\renewcommand{\arraystretch}{1.08}
\begin{tabular}{@{}lcccc@{}}
\toprule
Method & \makecell{PA-\\MPJPE$\downarrow$} & \makecell{PA-\\MPVPE$\downarrow$} & F@5$\uparrow$ & F@15$\uparrow$ \\
\midrule
% AMVUR\cite{jiang2023probabilistic}  & 8.7 & 8.3 & 0.593 & 0.964 \\
SPMHand\cite{lu2024spmhand} & 8.8 & 8.6 & 0.574 & 0.962 \\
MaskHand$^*$ \cite{saleem2025maskhand} & 7.0 & 7.0 & 0.663 & 0.984 \\
Hamba$^*$ \cite{dong2024hamba} & 6.9 & 6.8 & 0.681 & 0.982 \\
HandOS \cite{chen2025handos} & 8.4 & 8.4 & 0.584 & 0.962\\
HandOS$^*$ \cite{chen2025handos} & 6.8 & 6.7 & 0.688 & 0.983 \\
\midrule
GeoHand (Ours) & 7.4 & 7.1 & 0.654 & 0.980 \\
GeoHand$^*$ (Ours) & \textbf{6.7} & \textbf{6.4} & \textbf{0.697} & \textbf{0.987} \\
\bottomrule
\end{tabular}
\end{table}

\paragraph{Results on DexYCB.}
The advantages of GeoHand are especially pronounced on DexYCB, which features complex hand-object interactions and severe depth ambiguity. As reported in Table \ref{tab:dexycb}, GeoHand dramatically reduces the absolute errors, achieving an MPJPE of 7.4 mm and an MPVPE of 7.1 mm, vastly outperforming previous methods such as MaskHand (11.7 mm MPJPE) and SimpleHand (12.4 mm MPJPE). Furthermore, it establishes a new state of the art in Procrustes-aligned metrics with a PA-MPJPE of 5.0 mm and a PA-MPVPE of 4.7 mm. This significant margin verifies that while RGB-only transformers struggle to recover metric depth during object grasping, GeoHand's frozen monocular geometry prior effectively resolves global scale and root-relative depth directly from the image plane.

\paragraph{Results on HO3Dv3.}
HO3Dv3 presents the greatest challenge, with massive self-occlusion and truncation caused by manipulation. As demonstrated in Table \ref{tab:ho3d}, our mixed-data trained model, GeoHand$^{*}$, achieves top-tier results with a PA-MPJPE of 6.7 mm, a PA-MPVPE of 6.4 mm, and an F@5 score of 0.697. Notably, this improvement is not only due to better global geometry estimation. Once geometric priors are injected, KQIR further uses them to refine uncertain fingertip and inter-finger regions by querying geometry-aware features around projected joints. This combination allows our method to outperform contemporary massive-data models such as HandOS and Hamba.

\paragraph{Results on FreiHAND.}
As shown in Table \ref{tab:freihand}, GeoHand achieves highly competitive performance on FreiHAND with a PA-MPJPE of 5.1 mm and an F@15 score of 0.993. While it marginally trails HandOS \cite{chen2025handos} by 0.1 mm on PA-MPJPE, this behavior aligns directly with the dataset's composition. FreiHAND primarily consists of clean, single-hand images with minimal background clutter and no external object occlusion. In such scenarios, the explicit metric depth and global geometry clues provided by MoGe2 yield diminishing returns compared to highly occluded interactive scenes. HandOS, which tightly couples a generalized object detector with pose estimation within a strictly supervised end-to-end framework, specifically excels at extracting structural features on unoccluded single hands. Nevertheless, the metric indicates that our decoupled geometry priors still complement appearance features effectively, allowing GeoHand to match or outperform other recent robust architectures like WiLoR and HaMeR without specific bounding-box alignments.

\begin{figure*}[t!]
\centering
\includegraphics[width=\VisualizationFig,keepaspectratio]{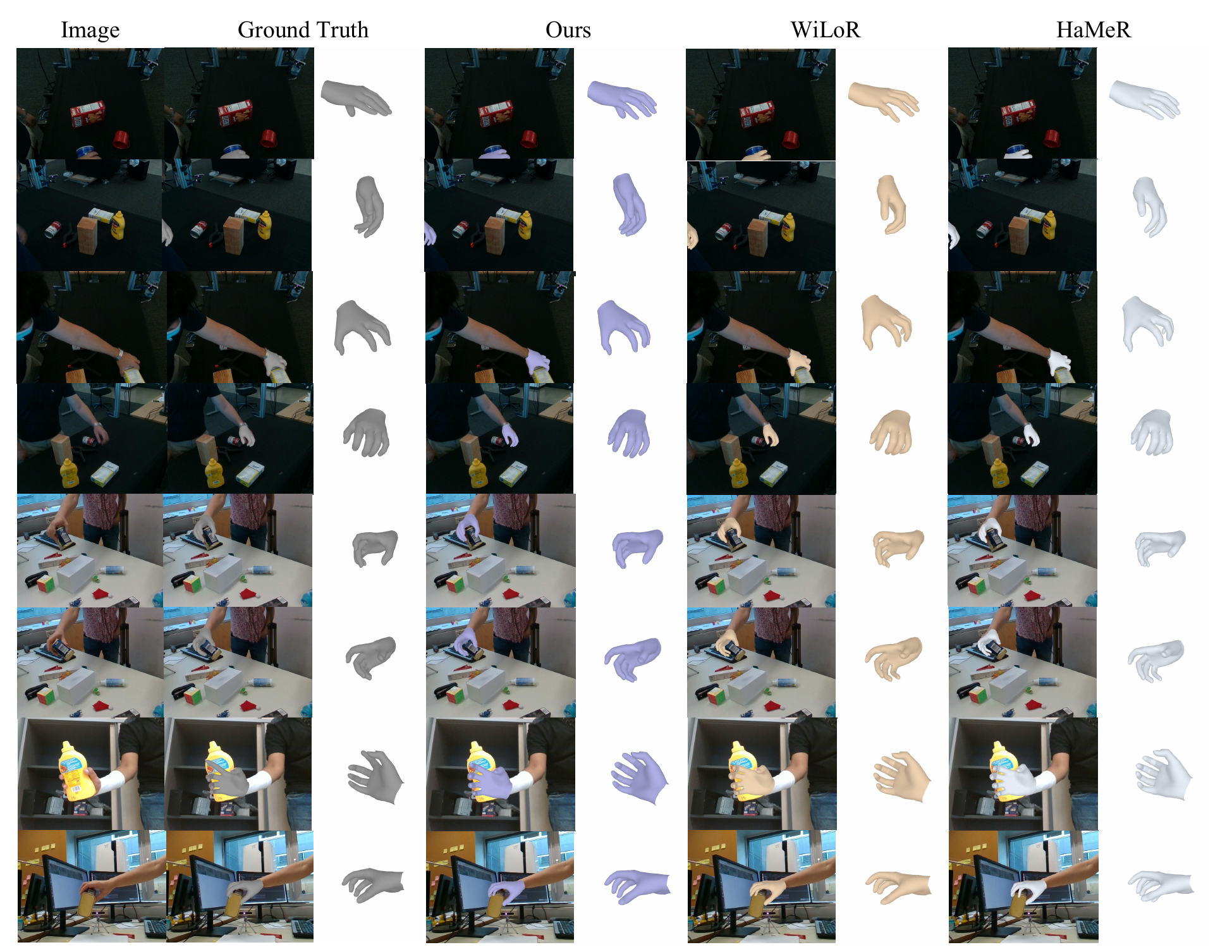}
\caption{Qualitative comparisons on DexYCB (top four rows) and HO3Dv3 (bottom four rows). Compared to WiLoR \cite{potamias2025wilor} and HaMeR \cite{pavlakos2024reconstructing}, GeoHand recovers precise finger articulations and maintains valid hand topologies under severe object occlusions.}
\label{fig:mesh_compare}
\end{figure*}

\begin{table}[t!]
\centering
\caption{Comparison on FreiHAND.}
\label{tab:freihand}
\setlength{\tabcolsep}{2.6pt}
\renewcommand{\arraystretch}{1.08}
\begin{tabular}{@{}lcccc@{}}
\toprule
Method & \makecell{PA-\\MPJPE$\downarrow$} & \makecell{PA-\\MPVPE$\downarrow$} & F@5$\uparrow$ & F@15$\uparrow$ \\
\midrule
MobRecon \cite{chen2022mobrecon} & 5.7 & 5.8 & 0.784 & 0.986 \\
SimpleHand \cite{zhou2024simple} & 5.7 & 6.0 & 0.772 & 0.986 \\
HaMeR \cite{pavlakos2024reconstructing} & 6.0 & 5.7 & 0.785 & 0.990 \\
WiLoR \cite{potamias2025wilor} & 5.5 & \textbf{5.1} & \textbf{0.825} & \textbf{0.993} \\
MaskHand \cite{saleem2025maskhand} & 5.5 & 5.4 & 0.801 & 0.991 \\
Hamba \cite{dong2024hamba} & 5.8 & 5.5 & 0.798 & 0.991 \\
HandOS \cite{chen2025handos} & \textbf{5.0} & 5.3 & 0.812 & 0.991 \\
\midrule
GeoHand (Ours) & 5.1 & 5.4 & 0.815 & \textbf{0.993} \\
\bottomrule
\end{tabular}
\end{table}

\subsection{Qualitative Results}

To visually demonstrate the superiority of our approach, we select HaMeR \cite{pavlakos2024reconstructing} and WiLoR \cite{potamias2025wilor} for visualization, as they are highly representative state-of-the-art transformer-based monocular frameworks with robust open-source weights. 

As shown in Figure \ref{fig:mesh_compare}, predicting 3D mesh articulation fundamentally relies on accurate geometry perception. The top rows present scenarios from DexYCB, while the bottom rows depict highly truncated interactions from HO3Dv3. In these complex object manipulation scenarios, RGB-only methods like HaMeR and WiLoR often struggle to infer correct spatial relationships. Due to severe self-occlusion and finger truncation introduced by grabbed objects, they occasionally produce significantly misaligned fingers or twisted hand meshes. In contrast, by unlocking and fusing high-fidelity prior geometry knowledge from the frozen MoGe2 architecture, GeoHand successfully maintains valid hand topologies and accurately recovers complex local finger articulations under severe object-induced truncations and occlusions.

\subsection{In-the-wild Generalization}
\label{sec:inthewild}

\begin{figure*}[t!]
\centering
\includegraphics[width=\VisualizationFigTwo,keepaspectratio]{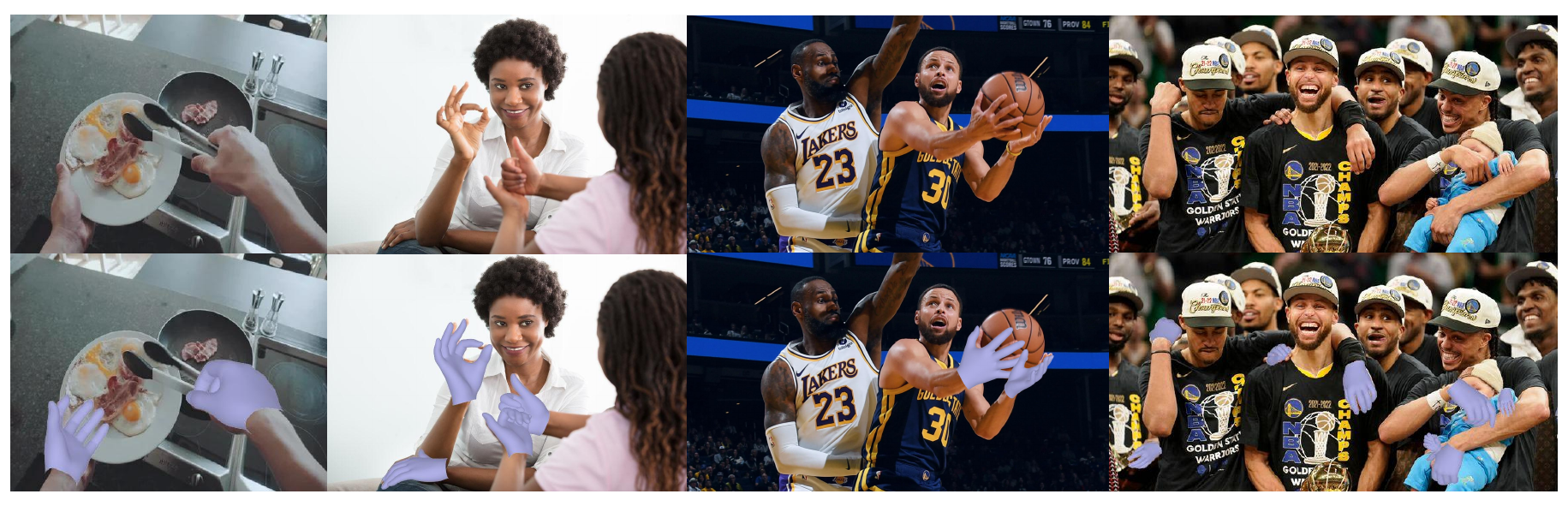}
\caption{Qualitative evaluation on in-the-wild multi-hand scenarios with crowded layouts and complex occlusions.}
\label{fig:realworld_multi}
\end{figure*}

To validate generalizability beyond curated benchmarks, we evaluate GeoHand on unconstrained in-the-wild images following recent protocols~\cite{chen2025handos, potamias2025wilor}. Real world scenarios naturally introduce complex backgrounds, arbitrary camera viewpoints, and diverse occlusions. Benefiting from the injected geometric priors, GeoHand maintains robust spatial awareness and successfully recovers accurate hand topologies in these challenging, unconstrained cases.

We demonstrate our strong performance in crowded multi-hand interactions (Figure~\ref{fig:realworld_multi}). GeoHand precisely resolves complex severe inter-hand overlaps from varied perspectives. Furthermore, qualitative evaluations on more diverse in-the-wild single hand images are documented in the Appendix. Crucially, these high-fidelity results are achieved strictly from a single RGB image, circumventing the need for explicit auxiliary modalities at inference time.

\subsection{Ablation Studies}

We perform ablation studies to validate the contribution of each key component in the GeoHand pipeline.

\begin{figure}[t!]
\centering
\includegraphics[width=1.0\linewidth]{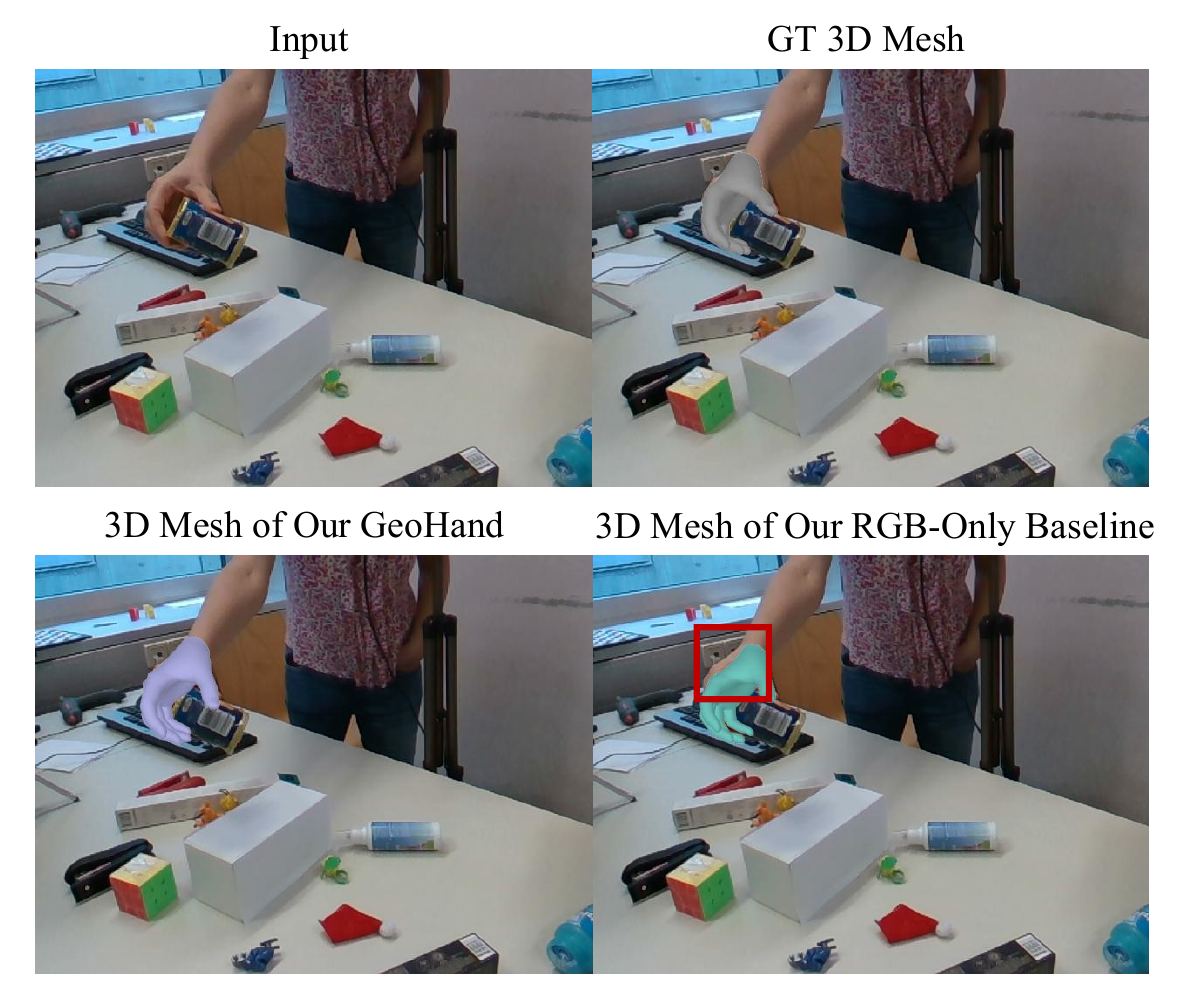}
\caption{Effect of geometric priors on HO3Dv3. GeoHand resolves the structural mismatch (red boxes) present in the RGB-only baseline.}
\label{fig:ablation_visual}
\end{figure}

\begin{figure}[t!]
\centering
\includegraphics[width=0.9\linewidth]{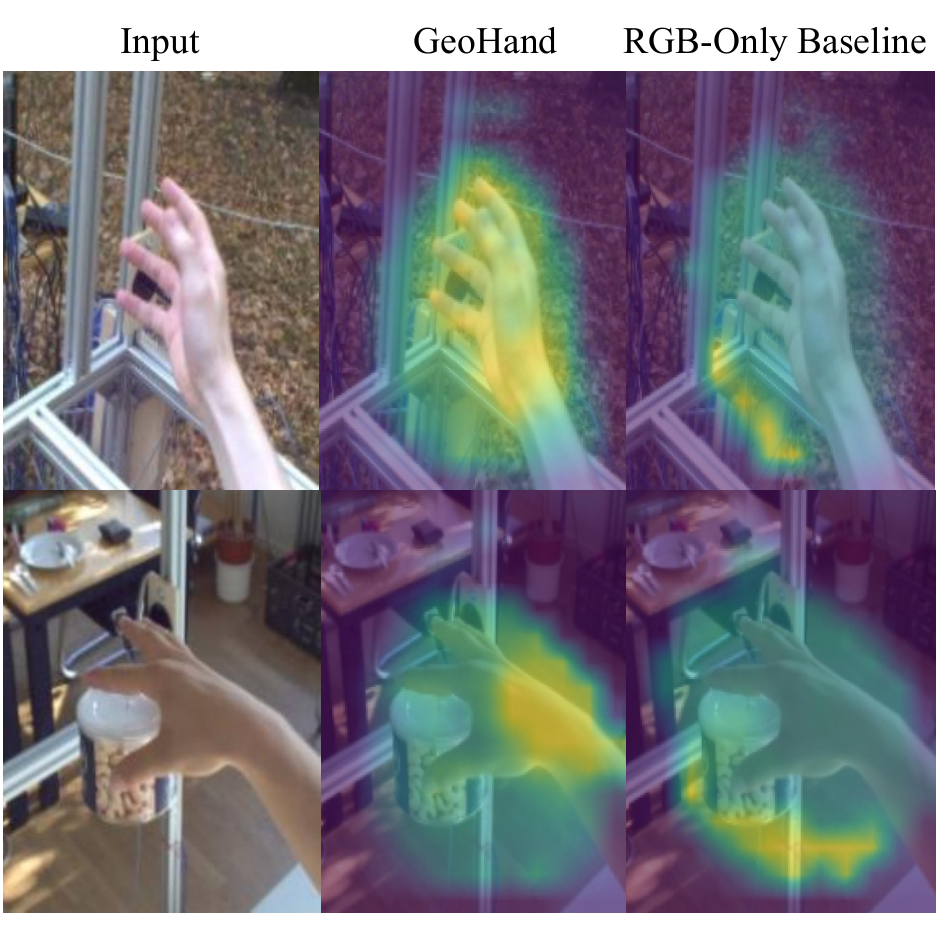}
\caption{Attention heatmaps on FreiHAND. Guided by geometric priors, GeoHand focuses entirely on hand structures, avoiding the background distractions seen in the RGB-only baseline.}
\label{fig:ablation_visual_heatmap}
\end{figure}

\paragraph{Visualization of geometric priors.}
To explicitly demonstrate the effectiveness of unlocking prior geometry knowledge, we present a visual comparison in Figure \ref{fig:ablation_visual} between our full GeoHand model and a pure RGB-only baseline. As highlighted by the red boxes, the RGB-only model fundamentally lacks spatial awareness, leaving a noticeable margin between the reconstructed mesh and the ground-truth contour. This is a common pitfall where models overfit to 2D texture boundaries and fail to perceive metric depth. In contrast, GeoHand, guided by high-fidelity MoGe2 priors, successfully bridges this gap. The explicit depth and normal embeddings constrain the 3D generation space, forcing the inferred mesh to align much more closely with the actual physical boundaries and perspective contours of the ground truth. Furthermore, attention heatmaps (Figure \ref{fig:ablation_visual_heatmap}) show geometric priors guide GeoHand to focus holistically on hand structures, eliminating the severe background distractions present in the RGB-only baseline.

\paragraph{Effect of geometry prior and GeoAdapter.}
Table \ref{tab:abl_geo_adapter} reveals that while raw MoGe2 tokens already provide a clear boost over the RGB-only baseline, integrating the map-level GeoAdapter yields further gains by explicitly aligning geometry semantics before tokenization. Raw geometry tokens extracted from foundational scene-level models often contain out-of-domain structures or background noise that can overwhelm the primary ViT attention maps if injected blindly. The GeoAdapter mitigates this by applying localized convolutional recalibration, filtering generic depth artifacts and aligning the geometric representation with human hands.

\begin{table}[t!] 
\centering
\caption{Ablation on geometry prior and GeoAdapter.}
\label{tab:abl_geo_adapter}
\setlength{\tabcolsep}{3pt}
\renewcommand{\arraystretch}{1.08}
\begin{tabular}{@{}lcc@{}}
\toprule
Configuration & \makecell{PA-\\MPJPE$\downarrow$} & \makecell{PA-\\MPVPE$\downarrow$} \\
\midrule
RGB-only baseline & 5.4 & 5.9 \\
+ MoGe2 tokens only & 5.2 & 5.6 \\
+ MoGe2 + GeoAdapter & \textbf{5.1} & \textbf{5.4} \\
\bottomrule
\end{tabular}
\end{table}

\paragraph{Efficacy and Efficiency of KQIR.}
To highlight the specific role of KQIR, we replace it with the refinement module used in WiLoR \cite{potamias2025wilor} and evaluate the variants on FreiHAND. As shown in Table \ref{tab:abl_refiner_comp}, substituting the WiLoR refiner with KQIR reduces PA-MPJPE from 5.4 mm to 5.1 mm while requiring less than half the parameters (1.125 M vs. 2.57 M). The maximum inference speed of KQIR (153 frames per second, FPS) remains highly competitive relative to the WiLoR refiner (156 FPS). This minor FPS drop, despite the large parameter reduction, is attributable to KQIR's dynamic query construction and spatially targeted cross-attention. While global mesh regressors incur massive computational overhead by simultaneously attending to all MANO vertices, our KQIR module focuses exclusively on the key articulated joints. By querying geometry-aware features strictly around projected 2D joint coordinates, it directly converts prior geometry into localized articulation corrections without redundant background processing.

\begin{table}[t!] 
\centering
\caption{Comparison of refinement modules on FreiHAND.}
\label{tab:abl_refiner_comp}
\setlength{\tabcolsep}{3pt}
\renewcommand{\arraystretch}{1.08}
\begin{tabular}{@{}lccc@{}}
\toprule
Refiner & Params$\downarrow$ & FPS$\uparrow$ & \makecell{PA-\\MPJPE$\downarrow$} \\
\midrule
WiLoR refiner \cite{potamias2025wilor} & 2.570 & \textbf{156} & 5.4 \\
KQIR, 2 steps (Ours) & \textbf{1.125} & 153 & \textbf{5.1} \\
\bottomrule
\end{tabular}
\end{table}

\paragraph{Effect of KQIR refinement steps.}
% Table \ref{tab:abl_KQIR} further measures the impact of iterative refinement depth. Iterative refinement is critical for resolving complex self-occlusions where single-pass predictions might drift. Two refinement steps reduce PA-MPJPE from 5.7 mm to 5.1 mm, confirming that structured querying around joint locations effectively exploits geometry-aware features. However, extending the process to three steps results in slight performance degradation (PA-MPJPE increases to 5.3 mm). We hypothesize this occurs because excessive refinement loops overfit to ambiguous texture bounds or noisy local geometry in heavily occluded areas, ultimately distorting the stable global kinematic chain established by the initial backbone.
Table \ref{tab:abl_KQIR} further evaluates the impact of KQIR refinement steps and their effect on inference speed. Two steps optimally reduce PA-MPJPE from 5.7 mm to 5.1 mm, effectively resolving self-occlusions where single-pass predictions drift. However, a third step degrades performance (5.3 mm), likely because excessive looping overfits to ambiguous local details, distorting the initially established global kinematic chain. Crucially, KQIR is extremely lightweight: the optimal two-step configuration maintains 153 FPS, which is only a marginal drop from the 160 FPS baseline. This demonstrates that KQIR significantly boosts accuracy with negligible impact on GeoHand's real-time efficiency.

\begin{table}[t!] 
\centering
\caption{Ablation on KQIR refinement steps.}
\label{tab:abl_KQIR}
\setlength{\tabcolsep}{2.4pt}
\renewcommand{\arraystretch}{1.08}
\begin{tabular}{@{}lccc@{}}
\toprule
Configuration & \makecell{PA-\\MPJPE$\downarrow$} & \makecell{PA-\\MPVPE$\downarrow$} & FPS$\uparrow$ \\
\midrule
Without KQIR & 5.7 & 6.1 & \textbf{160}\\ 
KQIR, 1 step & 5.4 & 5.8 & 156\\
KQIR, 2 steps & \textbf{5.1} & \textbf{5.4} & 153\\
KQIR, 3 steps & 5.3 & 5.7 & 149 \\
\bottomrule
\end{tabular}
\end{table}

\paragraph{Effect of bone-length and vertex losses.}
Table \ref{tab:abl_loss} validates our lightweight structural regularization. The explicit bone-length and vertex constraints effectively restrict the optimization space to anatomically valid solutions without relying on explicit adversarial discriminators. Without $\mathcal{L}_{bone}$, the network tends to improperly satisfy 2D reprojection constraints by unnaturally stretching or shrinking finger segments, thereby violating strict human biomechanical limits. The inclusion of $\mathcal{L}_{vert}$ further anchors the dense surface to the underlying skeletal topology, promoting smoother and more anatomically plausible skinning matrices.

\begin{table}[t!] 
\centering
\caption{Ablation on bone-length and vertex losses.}
\label{tab:abl_loss}
\setlength{\tabcolsep}{3pt}
\renewcommand{\arraystretch}{1.08}
\begin{tabular}{@{}lcc@{}}
\toprule
Configuration & \makecell{PA-\\MPJPE$\downarrow$} & \makecell{PA-\\MPVPE$\downarrow$} \\
\midrule
w/o $\mathcal{L}_{bone}$ & 5.3 & 5.7 \\
w/o $\mathcal{L}_{vert}$ & 5.3 & 5.7 \\
Full loss & \textbf{5.1} & \textbf{5.4} \\
\bottomrule
\end{tabular}
\end{table}

\section{Conclusion}
\label{sec:conclusion}

We presented GeoHand, a framework that leverages prior geometry knowledge for monocular 3D hand reconstruction. GeoHand enhances frozen MoGe2 geometry features with a map-level GeoAdapter before tokenization, fuses geometry into the backbone through a gated token fusion strategy, and further refines MANO predictions with the Keypoint-Queried Iterative Refiner. This design explicitly connects global geometric disambiguation with local articulation correction while preserving a pure RGB-only input pipeline. Extensive experiments demonstrate that GeoHand achieves state-of-the-art or highly competitive performance across diverse and challenging scenarios including the FreiHAND, DexYCB, and HO3Dv3 benchmarks. Notably, thanks to the robust structural priors, GeoHand shows exceptional resilience against severe self-occlusions and hand-object interactions, paving a promising direction for future robust 3D hand reconstruction systems.

\bibliographystyle{unsrtnat}
\bibliography{main}

\begin{thebibliography}{71}
\providecommand{\natexlab}[1]{#1}
\providecommand{\url}[1]{\texttt{#1}}
\expandafter\ifx\csname urlstyle\endcsname\relax
  \providecommand{\doi}[1]{doi: #1}\else
  \providecommand{\doi}{doi: \begingroup \urlstyle{rm}\Url}\fi

\bibitem[Zimmermann et~al.(2019)Zimmermann, Ceylan, Yang, Russell, Argus, and
  Brox]{zimmermann2019freihand}
Christian Zimmermann, Duygu Ceylan, Jimei Yang, Bryan Russell, Max Argus, and
  Thomas Brox.
\newblock Freihand: A dataset for markerless capture of hand pose and shape
  from single rgb images.
\newblock In \emph{Proceedings of the IEEE/CVF international conference on
  computer vision}, pages 813--822, 2019.

\bibitem[Chen et~al.(2022)Chen, Liu, Dong, Zhang, Ma, Xiong, Zhang, and
  Guo]{chen2022mobrecon}
Xingyu Chen, Yufeng Liu, Yajiao Dong, Xiong Zhang, Chongyang Ma, Yanmin Xiong,
  Yuan Zhang, and Xiaoyan Guo.
\newblock Mobrecon: Mobile-friendly hand mesh reconstruction from monocular
  image.
\newblock In \emph{Proceedings of the IEEE/CVF conference on computer vision
  and pattern recognition}, pages 20544--20554, 2022.

\bibitem[Pavlakos et~al.(2024)Pavlakos, Shan, Radosavovic, Kanazawa, Fouhey,
  and Malik]{pavlakos2024reconstructing}
Georgios Pavlakos, Dandan Shan, Ilija Radosavovic, Angjoo Kanazawa, David
  Fouhey, and Jitendra Malik.
\newblock Reconstructing hands in 3d with transformers.
\newblock In \emph{Proceedings of the IEEE/CVF Conference on Computer Vision
  and Pattern Recognition}, pages 9826--9836, 2024.

\bibitem[Pavlakos et~al.(2019)Pavlakos, Choutas, Ghorbani, Bolkart, Osman,
  Tzionas, and Black]{pavlakos2019expressive}
Georgios Pavlakos, Vasileios Choutas, Nima Ghorbani, Timo Bolkart, Ahmed~AA
  Osman, Dimitrios Tzionas, and Michael~J Black.
\newblock Expressive body capture: 3d hands, face, and body from a single
  image.
\newblock In \emph{Proceedings of the IEEE/CVF conference on computer vision
  and pattern recognition}, pages 10975--10985, 2019.

\bibitem[Romero et~al.(2022)Romero, Tzionas, and Black]{romero2022embodied}
Javier Romero, Dimitrios Tzionas, and Michael~J Black.
\newblock Embodied hands: Modeling and capturing hands and bodies together.
\newblock \emph{arXiv preprint arXiv:2201.02610}, 2022.

\bibitem[Bogo et~al.(2016)Bogo, Kanazawa, Lassner, Gehler, Romero, and
  Black]{bogo2016keep}
Federica Bogo, Angjoo Kanazawa, Christoph Lassner, Peter Gehler, Javier Romero,
  and Michael~J Black.
\newblock Keep it smpl: Automatic estimation of 3d human pose and shape from a
  single image.
\newblock In \emph{European conference on computer vision}, pages 561--578.
  Springer, 2016.

\bibitem[Kolotouros et~al.(2019{\natexlab{a}})Kolotouros, Pavlakos, Black, and
  Daniilidis]{kolotouros2019learning}
Nikos Kolotouros, Georgios Pavlakos, Michael~J Black, and Kostas Daniilidis.
\newblock Learning to reconstruct 3d human pose and shape via model-fitting in
  the loop.
\newblock In \emph{Proceedings of the IEEE/CVF international conference on
  computer vision}, pages 2252--2261, 2019{\natexlab{a}}.

\bibitem[Oberweger et~al.(2015)Oberweger, Wohlhart, and
  Lepetit]{oberweger2015hands}
Markus Oberweger, Paul Wohlhart, and Vincent Lepetit.
\newblock Hands deep in deep learning for hand pose estimation.
\newblock \emph{arXiv preprint arXiv:1502.06807}, 2015.

\bibitem[Boukhayma et~al.(2019)Boukhayma, Bem, and Torr]{boukhayma20193d}
Adnane Boukhayma, Rodrigo~de Bem, and Philip~HS Torr.
\newblock 3d hand shape and pose from images in the wild.
\newblock In \emph{Proceedings of the IEEE/CVF conference on computer vision
  and pattern recognition}, pages 10843--10852, 2019.

\bibitem[Zhang et~al.(2019)Zhang, Li, Mo, Zhang, and Zheng]{zhang2019end}
Xiong Zhang, Qiang Li, Hong Mo, Wenbo Zhang, and Wen Zheng.
\newblock End-to-end hand mesh recovery from a monocular rgb image.
\newblock In \emph{Proceedings of the IEEE/CVF international conference on
  computer vision}, pages 2354--2364, 2019.

\bibitem[Huang et~al.(2020)Huang, Ren, Wang, Qi, and Sun]{huang2020awr}
Weiting Huang, Pengfei Ren, Jingyu Wang, Qi~Qi, and Haifeng Sun.
\newblock Awr: Adaptive weighting regression for 3d hand pose estimation.
\newblock In \emph{Proceedings of the AAAI Conference on Artificial
  Intelligence}, volume~34, pages 11061--11068, 2020.

\bibitem[Mueller et~al.(2018)Mueller, Bernard, Sotnychenko, Mehta, Sridhar,
  Casas, and Theobalt]{mueller2018ganerated}
Franziska Mueller, Florian Bernard, Oleksandr Sotnychenko, Dushyant Mehta,
  Srinath Sridhar, Dan Casas, and Christian Theobalt.
\newblock Ganerated hands for real-time 3d hand tracking from monocular rgb.
\newblock In \emph{Proceedings of the IEEE conference on computer vision and
  pattern recognition}, pages 49--59, 2018.

\bibitem[Kulon et~al.(2020)Kulon, Guler, Kokkinos, Bronstein, and
  Zafeiriou]{kulon2020weakly}
Dominik Kulon, Riza~Alp Guler, Iasonas Kokkinos, Michael~M Bronstein, and
  Stefanos Zafeiriou.
\newblock Weakly-supervised mesh-convolutional hand reconstruction in the wild.
\newblock In \emph{Proceedings of the IEEE/CVF conference on computer vision
  and pattern recognition}, pages 4990--5000, 2020.

\bibitem[Baek et~al.(2019)Baek, Kim, and Kim]{baek2019pushing}
Seungryul Baek, Kwang~In Kim, and Tae-Kyun Kim.
\newblock Pushing the envelope for rgb-based dense 3d hand pose estimation via
  neural rendering.
\newblock In \emph{Proceedings of the IEEE/CVF conference on computer vision
  and pattern recognition}, pages 1067--1076, 2019.

\bibitem[Ge et~al.(2019)Ge, Ren, Li, Xue, Wang, Cai, and Yuan]{ge20193d}
Liuhao Ge, Zhou Ren, Yuncheng Li, Zehao Xue, Yingying Wang, Jianfei Cai, and
  Junsong Yuan.
\newblock 3d hand shape and pose estimation from a single rgb image.
\newblock In \emph{Proceedings of the IEEE/CVF conference on computer vision
  and pattern recognition}, pages 10833--10842, 2019.

\bibitem[Kolotouros et~al.(2019{\natexlab{b}})Kolotouros, Pavlakos, and
  Daniilidis]{kolotouros2019convolutional}
Nikos Kolotouros, Georgios Pavlakos, and Kostas Daniilidis.
\newblock Convolutional mesh regression for single-image human shape
  reconstruction.
\newblock In \emph{Proceedings of the IEEE/CVF conference on computer vision
  and pattern recognition}, pages 4501--4510, 2019{\natexlab{b}}.

\bibitem[Chen et~al.(2021)Chen, Liu, Ma, Chang, Wang, Chen, Guo, Wan, and
  Zheng]{chen2021camera}
Xingyu Chen, Yufeng Liu, Chongyang Ma, Jianlong Chang, Huayan Wang, Tian Chen,
  Xiaoyan Guo, Pengfei Wan, and Wen Zheng.
\newblock Camera-space hand mesh recovery via semantic aggregation and adaptive
  2d-1d registration.
\newblock In \emph{Proceedings of the IEEE/CVF conference on computer vision
  and pattern recognition}, pages 13274--13283, 2021.

\bibitem[Dosovitskiy et~al.(2020)Dosovitskiy, Beyer, Kolesnikov, Weissenborn,
  Zhai, Unterthiner, Dehghani, Minderer, Heigold, Gelly,
  et~al.]{dosovitskiy2020image}
Alexey Dosovitskiy, Lucas Beyer, Alexander Kolesnikov, Dirk Weissenborn,
  Xiaohua Zhai, Thomas Unterthiner, Mostafa Dehghani, Matthias Minderer, Georg
  Heigold, Sylvain Gelly, et~al.
\newblock An image is worth 16x16 words: Transformers for image recognition at
  scale.
\newblock \emph{arXiv preprint arXiv:2010.11929}, 2020.

\bibitem[He et~al.(2022)He, Chen, Xie, Li, Doll{\'a}r, and
  Girshick]{he2022masked}
Kaiming He, Xinlei Chen, Saining Xie, Yanghao Li, Piotr Doll{\'a}r, and Ross
  Girshick.
\newblock Masked autoencoders are scalable vision learners.
\newblock In \emph{Proceedings of the IEEE/CVF conference on computer vision
  and pattern recognition}, pages 16000--16009, 2022.

\bibitem[Liu et~al.(2021)Liu, Lin, Cao, Hu, Wei, Zhang, Lin, and
  Guo]{liu2021swin}
Ze~Liu, Yutong Lin, Yue Cao, Han Hu, Yixuan Wei, Zheng Zhang, Stephen Lin, and
  Baining Guo.
\newblock Swin transformer: Hierarchical vision transformer using shifted
  windows.
\newblock In \emph{Proceedings of the IEEE/CVF international conference on
  computer vision}, pages 10012--10022, 2021.

\bibitem[Potamias et~al.(2025)Potamias, Zhang, Deng, and
  Zafeiriou]{potamias2025wilor}
Rolandos~Alexandros Potamias, Jinglei Zhang, Jiankang Deng, and Stefanos
  Zafeiriou.
\newblock Wilor: End-to-end 3d hand localization and reconstruction
  in-the-wild.
\newblock In \emph{Proceedings of the Computer Vision and Pattern Recognition
  Conference}, pages 12242--12254, 2025.

\bibitem[Lin et~al.(2021{\natexlab{a}})Lin, Wang, and Liu]{lin2021end}
Kevin Lin, Lijuan Wang, and Zicheng Liu.
\newblock End-to-end human pose and mesh reconstruction with transformers.
\newblock In \emph{Proceedings of the IEEE/CVF conference on computer vision
  and pattern recognition}, pages 1954--1963, 2021{\natexlab{a}}.

\bibitem[Lin et~al.(2021{\natexlab{b}})Lin, Wang, and Liu]{lin2021mesh}
Kevin Lin, Lijuan Wang, and Zicheng Liu.
\newblock Mesh graphormer.
\newblock In \emph{Proceedings of the IEEE/CVF international conference on
  computer vision}, pages 12939--12948, 2021{\natexlab{b}}.

\bibitem[Cho et~al.(2022)Cho, Youwang, and Oh]{cho2022cross}
Junhyeong Cho, Kim Youwang, and Tae-Hyun Oh.
\newblock Cross-attention of disentangled modalities for 3d human mesh recovery
  with transformers.
\newblock In \emph{European Conference on Computer Vision}, pages 342--359.
  Springer, 2022.

\bibitem[Zhou et~al.(2024)Zhou, Zhou, Lv, Zou, Tang, and Liang]{zhou2024simple}
Zhishan Zhou, Shihao Zhou, Zhi Lv, Minqiang Zou, Yao Tang, and Jiajun Liang.
\newblock A simple baseline for efficient hand mesh reconstruction.
\newblock In \emph{Proceedings of the IEEE/CVF Conference on Computer Vision
  and Pattern Recognition}, pages 1367--1376, 2024.

\bibitem[Moon et~al.(2018)Moon, Chang, and Lee]{moon2018v2v}
Gyeongsik Moon, Ju~Yong Chang, and Kyoung~Mu Lee.
\newblock V2v-posenet: Voxel-to-voxel prediction network for accurate 3d hand
  and human pose estimation from a single depth map.
\newblock In \emph{Proceedings of the IEEE conference on computer vision and
  pattern Recognition}, pages 5079--5088, 2018.

\bibitem[Chao et~al.(2021)Chao, Yang, Xiang, Molchanov, Handa, Tremblay,
  Narang, Van~Wyk, Iqbal, Birchfield, et~al.]{chao2021dexycb}
Yu-Wei Chao, Wei Yang, Yu~Xiang, Pavlo Molchanov, Ankur Handa, Jonathan
  Tremblay, Yashraj~S Narang, Karl Van~Wyk, Umar Iqbal, Stan Birchfield, et~al.
\newblock Dexycb: A benchmark for capturing hand grasping of objects.
\newblock In \emph{Proceedings of the IEEE/CVF conference on computer vision
  and pattern recognition}, pages 9044--9053, 2021.

\bibitem[Hampali et~al.(2020)Hampali, Rad, Oberweger, and
  Lepetit]{hampali2020honnotate}
Shreyas Hampali, Mahdi Rad, Markus Oberweger, and Vincent Lepetit.
\newblock Honnotate: A method for 3d annotation of hand and object poses.
\newblock In \emph{Proceedings of the IEEE/CVF conference on computer vision
  and pattern recognition}, pages 3196--3206, 2020.

\bibitem[Taheri et~al.(2020)Taheri, Ghorbani, Black, and
  Tzionas]{taheri2020grab}
Omid Taheri, Nima Ghorbani, Michael~J Black, and Dimitrios Tzionas.
\newblock Grab: A dataset of whole-body human grasping of objects.
\newblock In \emph{European conference on computer vision}, pages 581--600.
  Springer, 2020.

\bibitem[Hasson et~al.(2019)Hasson, Varol, Tzionas, Kalevatykh, Black, Laptev,
  and Schmid]{hasson2019learning}
Yana Hasson, Gul Varol, Dimitrios Tzionas, Igor Kalevatykh, Michael~J Black,
  Ivan Laptev, and Cordelia Schmid.
\newblock Learning joint reconstruction of hands and manipulated objects.
\newblock In \emph{Proceedings of the IEEE/CVF conference on computer vision
  and pattern recognition}, pages 11807--11816, 2019.

\bibitem[Brahmbhatt et~al.(2020)Brahmbhatt, Tang, Twigg, Kemp, and
  Hays]{brahmbhatt2020contactpose}
Samarth Brahmbhatt, Chengcheng Tang, Christopher~D Twigg, Charles~C Kemp, and
  James Hays.
\newblock Contactpose: A dataset of grasps with object contact and hand pose.
\newblock In \emph{European Conference on Computer Vision}, pages 361--378.
  Springer, 2020.

\bibitem[Liu et~al.(2024)Liu, Ren, Gao, Wang, Sun, Qi, Zhuang, and
  Liao]{liu2024keypoint}
Xingyu Liu, Pengfei Ren, Yuanyuan Gao, Jingyu Wang, Haifeng Sun, Qi~Qi, Zirui
  Zhuang, and Jianxin Liao.
\newblock Keypoint fusion for rgb-d based 3d hand pose estimation.
\newblock In \emph{Proceedings of the AAAI Conference on Artificial
  Intelligence}, volume~38, pages 3756--3764, 2024.

\bibitem[Qi et~al.(2017{\natexlab{a}})Qi, Su, Mo, and Guibas]{qi2017pointnet}
Charles~R Qi, Hao Su, Kaichun Mo, and Leonidas~J Guibas.
\newblock Pointnet: Deep learning on point sets for 3d classification and
  segmentation.
\newblock In \emph{Proceedings of the IEEE conference on computer vision and
  pattern recognition}, pages 652--660, 2017{\natexlab{a}}.

\bibitem[Qi et~al.(2017{\natexlab{b}})Qi, Yi, Su, and Guibas]{qi2017pointnet++}
Charles~Ruizhongtai Qi, Li~Yi, Hao Su, and Leonidas~J Guibas.
\newblock Pointnet++: Deep hierarchical feature learning on point sets in a
  metric space.
\newblock \emph{Advances in neural information processing systems}, 30,
  2017{\natexlab{b}}.

\bibitem[Yang et~al.(2024)Yang, Kang, Huang, Xu, Feng, and Zhao]{yang2024depth}
Lihe Yang, Bingyi Kang, Zilong Huang, Xiaogang Xu, Jiashi Feng, and Hengshuang
  Zhao.
\newblock Depth anything: Unleashing the power of large-scale unlabeled data.
\newblock In \emph{Proceedings of the IEEE/CVF conference on computer vision
  and pattern recognition}, pages 10371--10381, 2024.

\bibitem[Wang et~al.(2025{\natexlab{a}})Wang, Xu, Dong, Deng, Xiang, Lv, Sun,
  Tong, and Yang]{wang2025moge}
Ruicheng Wang, Sicheng Xu, Yue Dong, Yu~Deng, Jianfeng Xiang, Zelong Lv,
  Guangzhong Sun, Xin Tong, and Jiaolong Yang.
\newblock Moge-2: Accurate monocular geometry with metric scale and sharp
  details.
\newblock \emph{arXiv preprint arXiv:2507.02546}, 2025{\natexlab{a}}.

\bibitem[Zimmermann and Brox(2017)]{zimmermann2017learning}
Christian Zimmermann and Thomas Brox.
\newblock Learning to estimate 3d hand pose from single rgb images.
\newblock In \emph{Proceedings of the IEEE international conference on computer
  vision}, pages 4903--4911, 2017.

\bibitem[Iqbal et~al.(2018)Iqbal, Molchanov, Gall, and Kautz]{iqbal2018hand}
Umar Iqbal, Pavlo Molchanov, Thomas Breuel~Juergen Gall, and Jan Kautz.
\newblock Hand pose estimation via latent 2.5 d heatmap regression.
\newblock In \emph{Proceedings of the European conference on computer vision
  (ECCV)}, pages 118--134, 2018.

\bibitem[Cai et~al.(2018)Cai, Ge, Cai, and Yuan]{cai2018weakly}
Yujun Cai, Liuhao Ge, Jianfei Cai, and Junsong Yuan.
\newblock Weakly-supervised 3d hand pose estimation from monocular rgb images.
\newblock In \emph{Proceedings of the European conference on computer vision
  (ECCV)}, pages 666--682, 2018.

\bibitem[Ge et~al.(2018)Ge, Cai, Weng, and Yuan]{ge2018hand}
Liuhao Ge, Yujun Cai, Junwu Weng, and Junsong Yuan.
\newblock Hand pointnet: 3d hand pose estimation using point sets.
\newblock In \emph{Proceedings of the IEEE conference on computer vision and
  pattern recognition}, pages 8417--8426, 2018.

\bibitem[Kanazawa et~al.(2018)Kanazawa, Black, Jacobs, and
  Malik]{kanazawa2018end}
Angjoo Kanazawa, Michael~J Black, David~W Jacobs, and Jitendra Malik.
\newblock End-to-end recovery of human shape and pose.
\newblock In \emph{Proceedings of the IEEE conference on computer vision and
  pattern recognition}, pages 7122--7131, 2018.

\bibitem[Rong et~al.(2021)Rong, Shiratori, and Joo]{rong2021frankmocap}
Yu~Rong, Takaaki Shiratori, and Hanbyul Joo.
\newblock Frankmocap: A monocular 3d whole-body pose estimation system via
  regression and integration.
\newblock In \emph{Proceedings of the IEEE/CVF International Conference on
  Computer Vision}, pages 1749--1759, 2021.

\bibitem[Tse et~al.(2022)Tse, Kim, Leonardis, and Chang]{tse2022collaborative}
Tze Ho~Elden Tse, Kwang~In Kim, Ales Leonardis, and Hyung~Jin Chang.
\newblock Collaborative learning for hand and object reconstruction with
  attention-guided graph convolution.
\newblock In \emph{Proceedings of the IEEE/CVF Conference on Computer Vision
  and Pattern Recognition}, pages 1664--1674, 2022.

\bibitem[Choi et~al.(2020)Choi, Moon, and Lee]{choi2020pose2mesh}
Hongsuk Choi, Gyeongsik Moon, and Kyoung~Mu Lee.
\newblock Pose2mesh: Graph convolutional network for 3d human pose and mesh
  recovery from a 2d human pose.
\newblock In \emph{European Conference on Computer Vision}, pages 769--787.
  Springer, 2020.

\bibitem[Yang and Yao(2019)]{yang2019disentangling}
Linlin Yang and Angela Yao.
\newblock Disentangling latent hands for image synthesis and pose estimation.
\newblock In \emph{Proceedings of the IEEE/CVF conference on computer vision
  and pattern recognition}, pages 9877--9886, 2019.

\bibitem[G{\"u}ler et~al.(2018)G{\"u}ler, Neverova, and
  Kokkinos]{guler2018densepose}
R{\i}za~Alp G{\"u}ler, Natalia Neverova, and Iasonas Kokkinos.
\newblock Densepose: Dense human pose estimation in the wild.
\newblock In \emph{Proceedings of the IEEE conference on computer vision and
  pattern recognition}, pages 7297--7306, 2018.

\bibitem[Moon and Lee(2020)]{moon2020i2l}
Gyeongsik Moon and Kyoung~Mu Lee.
\newblock I2l-meshnet: Image-to-lixel prediction network for accurate 3d human
  pose and mesh estimation from a single rgb image.
\newblock In \emph{European Conference on Computer Vision}, pages 752--768.
  Springer, 2020.

\bibitem[Kolotouros et~al.(2021)Kolotouros, Pavlakos, Jayaraman, and
  Daniilidis]{kolotouros2021probabilistic}
Nikos Kolotouros, Georgios Pavlakos, Dinesh Jayaraman, and Kostas Daniilidis.
\newblock Probabilistic modeling for human mesh recovery.
\newblock In \emph{Proceedings of the IEEE/CVF international conference on
  computer vision}, pages 11605--11614, 2021.

\bibitem[Park et~al.(2022)Park, Oh, Moon, Choi, and Lee]{park2022handoccnet}
JoonKyu Park, Yeonguk Oh, Gyeongsik Moon, Hongsuk Choi, and Kyoung~Mu Lee.
\newblock Handoccnet: Occlusion-robust 3d hand mesh estimation network.
\newblock In \emph{Proceedings of the IEEE/CVF conference on computer vision
  and pattern recognition}, pages 1496--1505, 2022.

\bibitem[Xu et~al.(2023)Xu, Wang, Tang, and Fu]{xu2023h2onet}
Hao Xu, Tianyu Wang, Xiao Tang, and Chi-Wing Fu.
\newblock H2onet: Hand-occlusion-and-orientation-aware network for real-time 3d
  hand mesh reconstruction.
\newblock In \emph{Proceedings of the IEEE/CVF conference on computer vision
  and pattern recognition}, pages 17048--17058, 2023.

\bibitem[Garcia-Hernando et~al.(2018)Garcia-Hernando, Yuan, Baek, and
  Kim]{garcia2018first}
Guillermo Garcia-Hernando, Shanxin Yuan, Seungryul Baek, and Tae-Kyun Kim.
\newblock First-person hand action benchmark with rgb-d videos and 3d hand pose
  annotations.
\newblock In \emph{Proceedings of the IEEE conference on computer vision and
  pattern recognition}, pages 409--419, 2018.

\bibitem[Moon et~al.(2020)Moon, Yu, Wen, Shiratori, and
  Lee]{moon2020interhand2}
Gyeongsik Moon, Shoou-I Yu, He~Wen, Takaaki Shiratori, and Kyoung~Mu Lee.
\newblock Interhand2. 6m: A dataset and baseline for 3d interacting hand pose
  estimation from a single rgb image.
\newblock In \emph{European Conference on Computer Vision}, pages 548--564.
  Springer, 2020.

\bibitem[Kwon et~al.(2021)Kwon, Tekin, St{\"u}hmer, Bogo, and
  Pollefeys]{kwon2021h2o}
Taein Kwon, Bugra Tekin, Jan St{\"u}hmer, Federica Bogo, and Marc Pollefeys.
\newblock H2o: Two hands manipulating objects for first person interaction
  recognition.
\newblock In \emph{Proceedings of the IEEE/CVF international conference on
  computer vision}, pages 10138--10148, 2021.

\bibitem[Yang et~al.(2022)Yang, Li, Zhan, Wu, Xu, Liu, and Lu]{yang2022oakink}
Lixin Yang, Kailin Li, Xinyu Zhan, Fei Wu, Anran Xu, Liu Liu, and Cewu Lu.
\newblock Oakink: A large-scale knowledge repository for understanding
  hand-object interaction.
\newblock In \emph{Proceedings of the IEEE/CVF conference on computer vision
  and pattern recognition}, pages 20953--20962, 2022.

\bibitem[Wang et~al.(2022)Wang, Zhang, Chen, Xu, Li, Liu, and
  Wang]{wang2022dexgraspnet}
Ruicheng Wang, Jialiang Zhang, Jiayi Chen, Yinzhen Xu, Puhao Li, Tengyu Liu,
  and He~Wang.
\newblock Dexgraspnet: A large-scale robotic dexterous grasp dataset for
  general objects based on simulation.
\newblock \emph{arXiv preprint arXiv:2210.02697}, 2022.

\bibitem[Grady et~al.(2021)Grady, Tang, Twigg, Vo, Brahmbhatt, and
  Kemp]{grady2021contactopt}
Patrick Grady, Chengcheng Tang, Christopher~D Twigg, Minh Vo, Samarth
  Brahmbhatt, and Charles~C Kemp.
\newblock Contactopt: Optimizing contact to improve grasps.
\newblock In \emph{Proceedings of the IEEE/CVF Conference on Computer Vision
  and Pattern Recognition}, pages 1471--1481, 2021.

\bibitem[Karunratanakul et~al.(2020)Karunratanakul, Spurr, Muigai, Hilliges,
  and Tang]{karunratanakul2020grasping}
Korrawe Karunratanakul, Adrian Spurr, Zicong Muigai, Otmar Hilliges, and Siyu
  Tang.
\newblock Grasping field: Learning implicit representations for human grasps.
\newblock In \emph{International Conference on 3D Vision (3DV)}, 2020.

\bibitem[Li et~al.(2022)Li, An, Zhang, Wu, Chen, Yu, and
  Liu]{li2022interacting}
Mengcheng Li, Liang An, Hongwen Zhang, Lianpeng Wu, Feng Chen, Tao Yu, and
  Yebin Liu.
\newblock Interacting attention graph for single image two-hand reconstruction.
\newblock In \emph{Proceedings of the IEEE/CVF conference on computer vision
  and pattern recognition}, pages 2761--2770, 2022.

\bibitem[Jiang et~al.(2024)Jiang, Zhou, Wang, Deng, Xu, and
  Shi]{jiang2024complementing}
Jianping Jiang, Xinyu Zhou, Bingxuan Wang, Xiaoming Deng, Chao Xu, and Boxin
  Shi.
\newblock Complementing event streams and rgb frames for hand mesh
  reconstruction.
\newblock In \emph{Proceedings of the IEEE/CVF Conference on Computer Vision
  and Pattern Recognition}, pages 24944--24954, 2024.

\bibitem[Spurr et~al.(2018)Spurr, Song, Park, and Hilliges]{spurr2018cross}
Adrian Spurr, Jie Song, Seonwook Park, and Otmar Hilliges.
\newblock Cross-modal deep variational hand pose estimation.
\newblock In \emph{Proceedings of the IEEE conference on computer vision and
  pattern recognition}, pages 89--98, 2018.

\bibitem[Ranftl et~al.(2020)Ranftl, Lasinger, Hafner, Schindler, and
  Koltun]{ranftl2020towards}
Ren{\'e} Ranftl, Katrin Lasinger, David Hafner, Konrad Schindler, and Vladlen
  Koltun.
\newblock Towards robust monocular depth estimation: Mixing datasets for
  zero-shot cross-dataset transfer.
\newblock \emph{IEEE transactions on pattern analysis and machine
  intelligence}, 44\penalty0 (3):\penalty0 1623--1637, 2020.

\bibitem[Ke et~al.(2024)Ke, Obukhov, Huang, Metzger, Daudt, and
  Schindler]{ke2024repurposing}
Bingxin Ke, Anton Obukhov, Shengyu Huang, Nando Metzger, Rodrigo~Caye Daudt,
  and Konrad Schindler.
\newblock Repurposing diffusion-based image generators for monocular depth
  estimation.
\newblock In \emph{Proceedings of the IEEE/CVF conference on computer vision
  and pattern recognition}, pages 9492--9502, 2024.

\bibitem[Wang et~al.(2025{\natexlab{b}})Wang, Xu, Dai, Xiang, Deng, Tong, and
  Yang]{wang2024moge}
Ruicheng Wang, Sicheng Xu, Cassie Dai, Jianfeng Xiang, Yu~Deng, Xin Tong, and
  Jiaolong Yang.
\newblock Moge: Unlocking accurate monocular geometry estimation for
  open-domain images with optimal training supervision.
\newblock In \emph{Proceedings of the IEEE/CVF Conference on Computer Vision
  and Pattern Recognition}, pages 5261--5271, 2025{\natexlab{b}}.

\bibitem[Zimmermann et~al.(2021)Zimmermann, Argus, and
  Brox]{zimmermann2021contrastive}
Christian Zimmermann, Max Argus, and Thomas Brox.
\newblock Contrastive representation learning for hand shape estimation.
\newblock In \emph{Proceedings of the DAGM German Conference on Pattern
  Recognition}, pages 250--264, 2021.

\bibitem[Yang et~al.(2020)Yang, Li, Xu, Diao, and Lu]{yang2020bihand}
Lixin Yang, Jiasen Li, Wenqiang Xu, Yiqun Diao, and Cewu Lu.
\newblock Bihand: Recovering hand mesh with multi-stage bisected hourglass
  networks.
\newblock \emph{arXiv preprint arXiv:2008.05079}, 2020.

\bibitem[Loshchilov and Hutter(2017)]{loshchilov2017decoupled}
Ilya Loshchilov and Frank Hutter.
\newblock Decoupled weight decay regularization.
\newblock \emph{arXiv preprint arXiv:1711.05101}, 2017.

\bibitem[Paszke et~al.(2019)Paszke, Gross, Massa, Lerer, Bradbury, Chanan,
  Killeen, Lin, Gimelshein, Antiga, et~al.]{paszke2019pytorch}
Adam Paszke, Sam Gross, Francisco Massa, Adam Lerer, James Bradbury, Gregory
  Chanan, Trevor Killeen, Zeming Lin, Natalia Gimelshein, Luca Antiga, et~al.
\newblock Pytorch: An imperative style, high-performance deep learning library.
\newblock \emph{Advances in neural information processing systems}, 32, 2019.

\bibitem[Dong et~al.(2024)Dong, Chharia, Gou, Carrasco, and De~la
  Torre]{dong2024hamba}
Haoye Dong, Aviral Chharia, Wenbo Gou, Francisco~Vicente Carrasco, and Fernando
  De~la Torre.
\newblock Hamba: Single-view 3d hand reconstruction with graph-guided
  bi-scanning mamba.
\newblock \emph{arXiv preprint arXiv:2407.09646}, 2024.

\bibitem[Chen et~al.(2025)Chen, Song, Jiang, Hu, Yu, and Zhang]{chen2025handos}
Xingyu Chen, Zhuheng Song, Xiaoke Jiang, Yaoqing Hu, Junzhi Yu, and Lei Zhang.
\newblock Handos: 3d hand reconstruction in one stage.
\newblock In \emph{Proceedings of the Computer Vision and Pattern Recognition
  Conference}, pages 17304--17314, 2025.

\bibitem[Saleem et~al.(2025)Saleem, Pinyoanuntapong, Patel, Xue, Helmy, Das,
  and Wang]{saleem2025maskhand}
Muhammad~Usama Saleem, Ekkasit Pinyoanuntapong, Mayur~Jagdishbhai Patel,
  Hongfei Xue, Ahmed Helmy, Srijan Das, and Pu~Wang.
\newblock Maskhand: Generative masked modeling for robust hand mesh
  reconstruction in the wild.
\newblock In \emph{Proceedings of the IEEE/CVF International Conference on
  Computer Vision}, pages 8372--8383, 2025.

\bibitem[Lu et~al.(2024)Lu, Gou, and Li]{lu2024spmhand}
Haofan Lu, Shuiping Gou, and Ruimin Li.
\newblock Spmhand: Segmentation-guided progressive multi-path 3d hand pose and
  shape estimation.
\newblock \emph{IEEE Transactions on Multimedia}, 26:\penalty0 6822--6833,
  2024.

\end{thebibliography}

\clearpage
\appendix
\section*{Appendix}
\setcounter{section}{1}
\setcounter{subsection}{0}
\subsection{Detailed Training Objectives}
\label{sec:appendix_loss}

GeoHand is optimized using a physically motivated combination of 2D, 3D, and MANO parametric loss functions. For a mini-batch of size $B$, let $N=21$ denote the number of joints, $V=778$ the mesh vertices, and $E=20$ the predefined bone segments. We formulate the final training objective as:
\begin{equation}
\begin{aligned}
\mathcal{L}_{total} &= 
1.0\,\mathcal{L}_{2D} 
+ 5.0\,\mathcal{L}_{3D\_joint} 
+ 1.0\,\mathcal{L}_{bone} \\
&\quad+ 0.1\,\mathcal{L}_{vert} 
+ 0.1\,\mathcal{L}_{global} 
+ 0.1\,\mathcal{L}_{pose} \\
&\quad+ 0.01\,\mathcal{L}_{betas} 
+ 0.05\,\mathcal{L}_{shape}.
\end{aligned}
\end{equation}

\paragraph{2D and 3D Structural Supervision.}
\begin{itemize}[leftmargin=1.5em]
    \item \textbf{2D Reprojection Loss $\mathcal{L}_{2D}$}: Computes the $L_1$ distance between the projected normalized 2D coordinates $\hat{\mathbf{u}}_{b,n}$ and their ground truth (GT) $\mathbf{u}_{b,n}$, masked by valid flags $m^{uv}_{b,n}$:
    \begin{equation}
    \mathcal{L}_{2D} = \frac{1}{B} \sum_{b=1}^{B} \frac{\sum_{n=1}^{N} m^{uv}_{b,n} \|\hat{\mathbf{u}}_{b,n} - \mathbf{u}_{b,n}\|_1}{\max(\sum_{n} m^{uv}_{b,n}, 1)}.
    \end{equation}

    \item \textbf{3D Joint Loss $\mathcal{L}_{3D\_joint}$}: Applies an $L_1$ loss on root-relative 3D joint coordinates, where $\tilde{\hat{\mathbf{X}}}_{b,n}$ and $\tilde{\mathbf{X}}_{b,n}$ denote the predicted and GT relative positions of joint $n$, respectively. To emphasize intricate terminal articulations, a higher spatial weight $w^{tip}_n=2.5$ is assigned to the five fingertips, while keeping $w_n=1.0$ for others. $m^{xyz}_b$ denotes the 3D validity flag.
    \begin{equation}
    \mathcal{L}_{3D\_joint} = \frac{1}{B} \sum_{b=1}^{B} \frac{\sum_{n=1}^{N} m^{xyz}_b \, w^{tip}_n \, \|\tilde{\hat{\mathbf{X}}}_{b,n} - \tilde{\mathbf{X}}_{b,n}\|_1}{\max(N m^{xyz}_b, 1)}.
    \end{equation}

    \item \textbf{Bone-Length Loss $\mathcal{L}_{bone}$}: Constrains severe anatomical distortions by penalizing the absolute error between the predicted bone length $\hat{d}_{b,e}$ and the GT bone length $d_{b,e}$ for each connected joint pair $(i_e, j_e)$:
    \begin{equation}
    \mathcal{L}_{bone} = \frac{1}{B} \sum_{b=1}^{B} \frac{\sum_{e=1}^{E} m^{xyz}_{b} |\hat{d}_{b,e} - d_{b,e}|}{\max(E m^{xyz}_{b}, 1)}.
    \end{equation}

    \item \textbf{Vertex Loss $\mathcal{L}_{vert}$}: Minimizes the dense surface discrepancy. Let $\hat{\mathbf{V}}_{b,v}$ and $\mathbf{V}_{b,v}$ denote the predicted and GT 3D coordinates for vertex $v$. To decouple hand shape from global translation, vertices are aligned using their respective predicted $\hat{\mathbf{R}}_b$ and GT $\mathbf{R}_b$ root joint positions. During training, GT vertices are reconstructed directly from GT MANO parameters to ensure parametric consistency. 
    \begin{equation}
    \mathcal{L}_{vert} = \frac{1}{3BV} \sum_{b=1}^{B} \sum_{v=1}^{V} \|(\hat{\mathbf{V}}_{b,v} - \hat{\mathbf{R}}_b) - (\mathbf{V}_{b,v} - \mathbf{R}_b)\|_1.
    \end{equation}
\end{itemize}

\paragraph{MANO Parameter Supervision.}
To enforce parametric fidelity, we apply a Smooth $L_1$ loss (denoted as $\rho_{\delta}$ with $\delta=0.05$) to the global orientation $\mathbf{G}$, hand pose $\mathbf{P}$ (both represented as flattened rotation matrices), and shape parameters $\boldsymbol{\beta}$. Let $\hat{\mathbf{G}}_{b,d}$ and $\mathbf{G}_{b,d}$ represent the $d$-th element of the predicted and GT flattened global orientation matrices. These losses are truncated dynamically to a maximum value of 1.0 per sample to prevent unstable gradients:
\begin{equation}
\mathcal{L}_{global} = \frac{1}{B} \sum_{b=1}^{B} \min \left( \frac{1}{9} \sum_{d=1}^{9} \rho_{0.05}(\hat{\mathbf{G}}_{b,d} - \mathbf{G}_{b,d}),\, 1.0 \right).
\end{equation}
The hand pose loss $\mathcal{L}_{pose}$ and shape parameter loss $\mathcal{L}_{betas}$ follow the exact same truncated formulation over their 135 and 10 dimensions, comparing predictions ($\hat{\mathbf{P}}$, $\hat{\boldsymbol{\beta}}$) against their GT counterparts ($\mathbf{P}$, $\boldsymbol{\beta}$).

\paragraph{Shape Regularization.}
Instead of the squared norm, an un-squared $L_2$ norm penalty $\mathcal{L}_{shape} = \frac{1}{B} \sum_{b=1}^{B} \|\hat{\boldsymbol{\beta}}_b\|_2$ is enforced on the predicted shape parameters $\hat{\boldsymbol{\beta}}_b$ to prevent large deviations from the zero-mean MANO shape prior.

\subsection{Evaluation Metrics Definition}
\label{sec:appendix_metrics}

We evaluated our methodology using a widely standardized set of metrics for 3D human and hand pose estimation tasks. Let $N$ represent the number of evaluated coordinates.

\begin{itemize}[leftmargin=1.5em]
    \item \textbf{MPJPE (Mean Per Joint Position Error):} Calculates the mean Euclidean distance (in mm) between estimated and ground-truth 3D joint coordinates after aligning the root joints to the origin:
    \begin{equation}
    \mathrm{MPJPE}
    = \frac{1}{N}\sum_{i=1}^{N}\|\mathbf{J}_i-\mathbf{J}_i^{gt}\|_2.
    \end{equation}

    \item \textbf{PA-MPJPE (Procrustes-Aligned MPJPE):} Employs Generalized Procrustes Analysis (GPA) to optimally align the predicted 3D joints to the ground truth joints in terms of translation, rotation, and uniform scale before calculating MPJPE. This metric primarily evaluates the precision of articulated shape and pose independent of global depth displacement errors.

    \item \textbf{MPVPE \& PA-MPVPE (Mean Per Vertex Position Error):} The high-density counterparts to joint errors, applying identical alignment protocols (root-aligned or Procrustes-aligned) directly across all 778 vertices of a MANO mesh surface.

    \item \textbf{F-score (F@5, F@15):}
    A harmonic mean combining precision and recall to evaluate the geometric proximity between the reconstructed mesh and the ground-truth surface. It measures the fraction of accurately estimated mesh vertices under strict physical distance tolerances (typically evaluated at thresholds of 5 mm and 15 mm).
\end{itemize}

\subsection{Additional Qualitative Results}
\label{sec:appendix_results}

\begin{figure*}[p]
    \centering
    \vspace*{\fill}
    \includegraphics[width=0.96\linewidth,height=0.30\textheight,keepaspectratio]{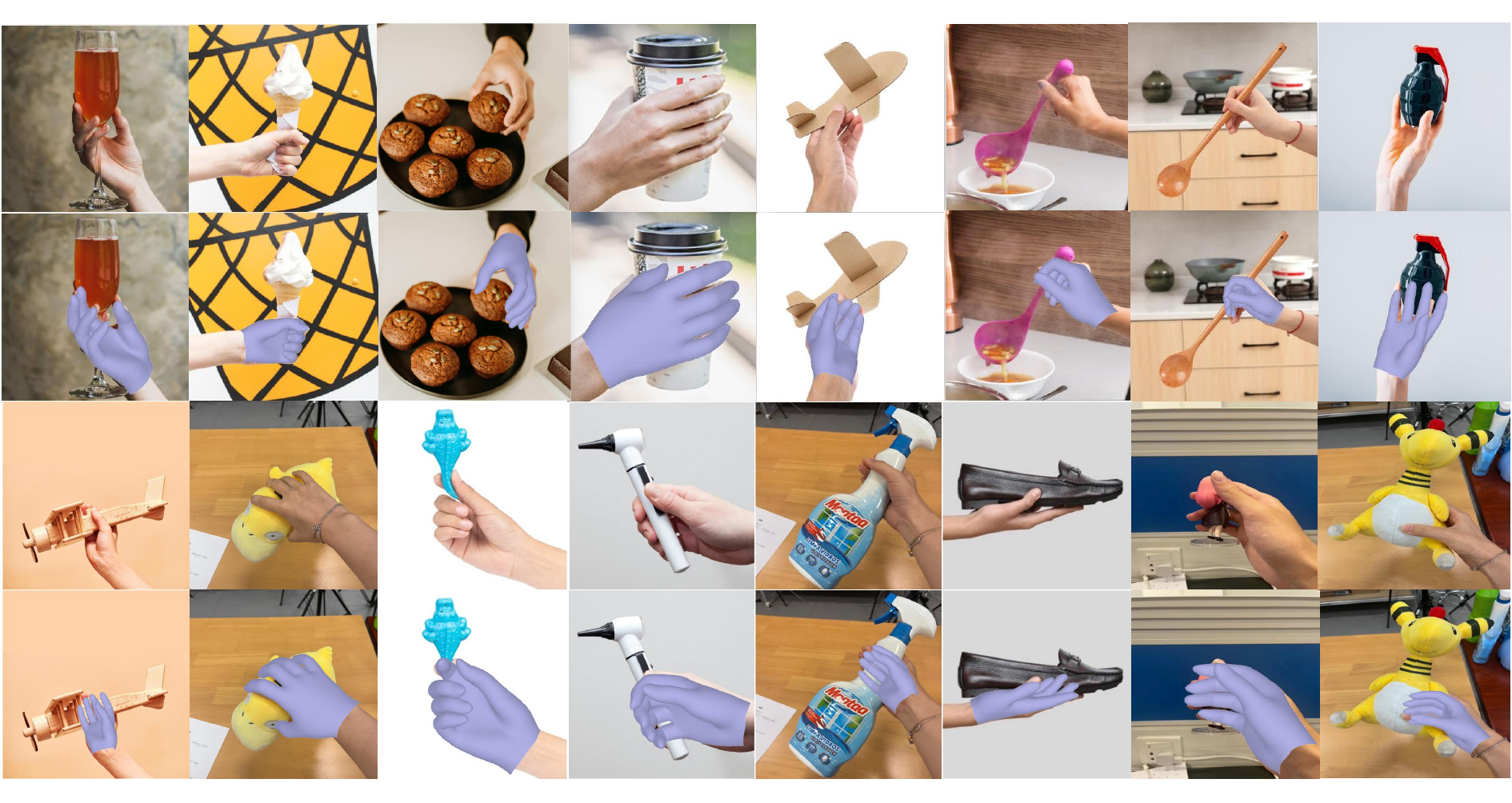}
    \caption{Qualitative evaluation of GeoHand in in-the-wild single-hand object interaction scenarios.}
    \label{fig:app_realworld1}
    \vspace{0pt}
    \includegraphics[width=1.0\linewidth,height=0.55\textheight,keepaspectratio]{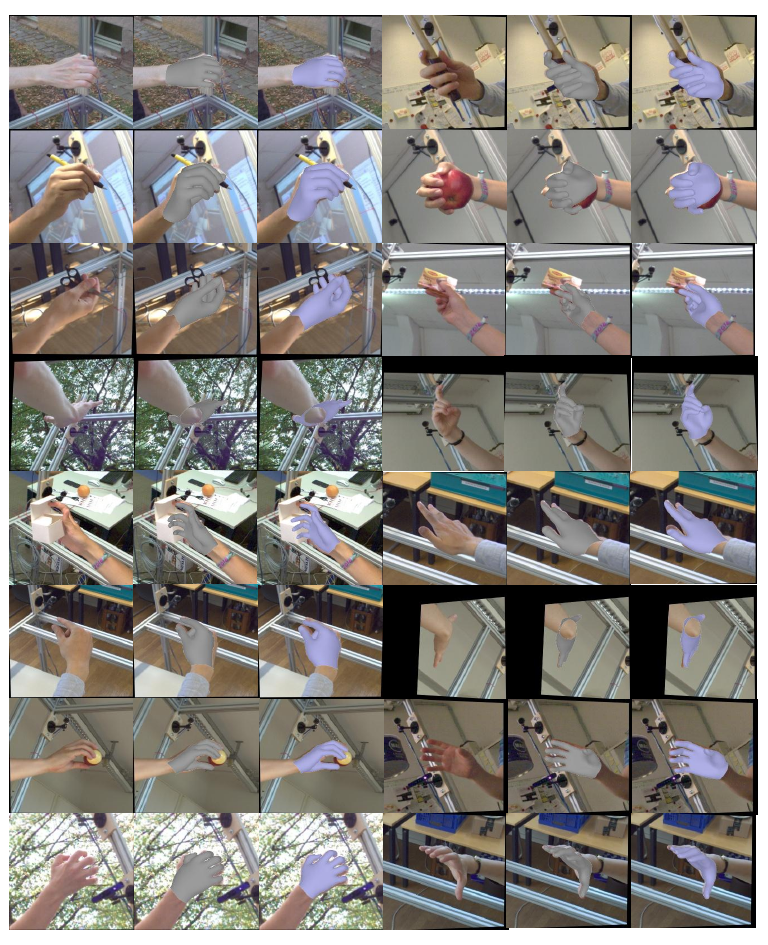}
    \caption{Visualization of the FreiHAND evaluation set. Samples in each triplet are the input image, the ground-truth mesh, and the mesh reconstructed by GeoHand.}
    \label{fig:app_freihand}
    \vspace*{\fill}
\end{figure*}

\paragraph{In-the-wild Single-Hand Scenarios.}
To thoroughly demonstrate generalization performance, we visualize single-hand object manipulations acquired directly from unconstrained in-the-wild scenarios, as detailed in Figure~\ref{fig:app_realworld1}. Despite severe arbitrary external truncations introduced by unknown objects and diverse backgrounds, GeoHand's reliance on integrated geometric embeddings ensures reliable joint prediction without mesh tearing.

\paragraph{Visualization on FreiHAND.}
Figure~\ref{fig:app_freihand} further illustrates the final 3D inferences on the FreiHAND evaluation set, showcasing the synergistic impact of GeoHand's architectural innovations. The robust overall mesh topology highlights how the injected geometric priors effectively resolve root-relative depth ambiguities and prevent coarse structural distortions. Building upon this globally consistent layout, the Keypoint-Queried Iterative Refiner (KQIR) leverages these enriched feature representations for targeted local corrections. By explicitly querying the geometry-aware maps around projected joint coordinates, KQIR precisely rectifies complex terminal articulations, such as overlapping fingertips. This cohesive interplay—where cross-modal fusion secures the global anatomical structure and KQIR refines local geometric details—enables GeoHand to consistently recover high-fidelity meshes from a single RGB image.

\subsection{Limitations and Future Work}
\label{sec:appendix_limitations}

While GeoHand significantly mitigates RGB-based spatial ambiguities by unlocking general monocular depth estimators, three primary directions are identified for extended optimization:

\begin{enumerate}[leftmargin=1.5em]
    \item \textbf{Exploration of Cross-Modal Extraction Limits.}
    Although the simple token concatenation and map-level GeoAdapter mechanism have proven highly effective against domain gaps without complex volume-rendering overheads, explicitly exploring more sophisticated multimodal Transformer frameworks with cross-attention between appearance tokens and coordinate-bound geometry matrices may unlock further precision gains.

    \item \textbf{Bounding-Box Reliance vs. End-to-End Synergy.}
    GeoHand functions under a top-down paradigm dependent on an independent 2D hand detector's bounding-box cropping stage. Consequently, final 3D reconstruction quality remains partially bounded by this preprocessing stage. Tightly unifying high-quality object detection within the same ViT-based pipeline would be a prompt and meaningful upgrade.

    \item \textbf{Temporal \& Multi-View Contextual Recovery for Occlusions.}
    While geometry priors substantially aid occluded interactions, recovering radically invisible topological boundaries remains inherently mathematically impossible from a static single viewpoint alone. Utilizing sequential historical motion context (temporal kinematic pipelines) or complementary multi-view spatial observations remains a promising direction toward high-precision immersive reconstruction.
\end{enumerate}

\clearpage

\end{document}